\PassOptionsToPackage{table}{xcolor}
\documentclass{article}

\usepackage{iclr2026/iclr2026_conference,times}
\iclrfinalcopy

\usepackage[T1]{fontenc}
\usepackage{microtype}
\microtypesetup{expansion=false}
\usepackage{amsmath,amssymb}
\usepackage{etoolbox}
\usepackage{graphicx}
\usepackage{subcaption}
\usepackage{booktabs}
\usepackage{multirow}
\usepackage{array}
\usepackage{placeins}
\usepackage[most]{tcolorbox}
\usepackage{hyperref}
\usepackage{url}

\definecolor{paperblue}{HTML}{2E5AA8}
\hypersetup{
  colorlinks=true,
  linkcolor=paperblue,
  citecolor=paperblue,
  urlcolor=paperblue,
  pdftitle={Sample-Efficient Learning from Agent Experience},
  pdfauthor={Chenhui Gou, Haoqin Tu, Yunhao Fang, Jianfei Cai, Hamid Rezatofighi}
}
\AtBeginEnvironment{equation}{\small}

\newcommand{\methodname}{Experience Distillation}
\newcommand{\methodabbr}{EPD}

\title{\makebox[\textwidth][c]{%
  \makebox[1.08\textwidth][c]{Sample-Efficient Learning from Agent Experience}%
}}

\author{
\begin{tabular}{c}
Chenhui Gou\textsuperscript{1,2,*} \enspace
Haoqin Tu\textsuperscript{2,\S} \enspace
Yunhao Fang\textsuperscript{2,\S} \enspace
Jianfei Cai\textsuperscript{1} \enspace
Hamid Rezatofighi\textsuperscript{1} \\[0.5ex]
{\normalfont\textsuperscript{1}Monash University \quad
\textsuperscript{2}ByteDance Seed}
\end{tabular}
}

\begin{document}
\maketitle
\pagestyle{plain}
\begingroup
\renewcommand{\thefootnote}{\fnsymbol{footnote}}
\footnotetext[1]{Correspondence: \texttt{Chenhui.Gou@monash.edu}.}
\footnotetext[4]{Work done while at ByteDance Seed.}
\endgroup

\begin{abstract}
Real-world agent learning is often constrained by costly environment interactions, such as running time-consuming experiments or obtaining human feedback. In-context learning offers a highly sample-efficient way for agents to learn from their own interaction histories, but its gains disappear once that experience is removed from the context. Separately, context distillation provides a mechanism for internalizing contextual information into model weights. However, applying it to agents' interaction histories without sacrificing environment sample efficiency remains underexplored. We term this problem \methodname{} and develop an implementation that requires no further environment interaction beyond the collected experience. Experiments on 749 curated software-engineering tasks and six text-adventure games show that it retains at least 64.8\% of the gains from in-context learning across both domains, whereas direct supervised fine-tuning on the collected experience recovers only 3.8\%. Compared with classical reinforcement-learning baselines, in-context learning from trial-and-error experience followed by \methodname{} matches their performance with at least \(9.6\times\) fewer environment samples.
\end{abstract}

\section{Introduction}
Real-world tasks can demand substantial time, resources, or human expertise. Testing a proposed battery material may require laboratory synthesis and measurement, whereas evaluating a legal analysis may require review by a qualified lawyer. As agents become capable of handling longer and more realistic tasks \citep{task-completion-time-horizons-of-frontier-ai-models,kwa2025measuring}, collecting rollouts from the environment becomes increasingly costly. Reinforcement learning (RL) is a standard framework for learning through interaction with an environment. However, it often has poor environment sample efficiency, requiring large numbers of environment interactions to achieve strong performance \citep{haarnoja2018soft,kaiser2020modelbased,narvekar2020curriculum}. This demand is manageable when environment interaction is cheap and scalable, such as in Go self-play \citep{silver2016mastering} or lightweight coding tasks \citep{guo2025deepseek}. When environment interaction is costly, the same sample demand can make training impractical.

In-context learning (ICL) offers a sample-efficient way for large language models to adapt using task-specific evidence in the context. By conditioning on a few demonstrations or on feedback from previous attempts, ICL can improve an agent's behavior without requiring a large training set or extensive environment interaction \citep{brown2020language,garg2022what,laskin2023incontext,zhu2026edgebench}. However, ICL does not update the model parameters, so the improvement disappears once the context is removed. Separately, context distillation provides a natural mechanism for consolidating these gains into model weights. It trains a student without informative context to reproduce the behavior of a teacher conditioned on that context \citep{hinton2015distilling,askell2021general,snell2022learning}. Combining ICL with context distillation therefore appears to offer a sample-efficient way to learn through interaction with an environment \citep{duan2024incontext}. Recent work has begun to explore this idea in agent tasks \citep{ye2026online,ye2026opcd,penaloza2026privileged,wang2026skillsd}. However, distilling an experience-conditioned teacher requires running it in the environment again. A practical, sample-efficient method should therefore minimize additional environment interaction during distillation. We term this problem \textbf{\methodname{}}.

To avoid further environment interaction, a naive solution inspired by model-based RL is to run the experience-conditioned teacher in a learned world model rather than the real environment \citep{sutton2018reinforcement,chua2018deep}. However, even small errors from the world model build up over long rollouts, making the generated trajectories unreliable \citep{talvitie2017self}. Branched rollouts reduce this problem by starting from points sampled along collected trajectories and rolling out in the world model for only a few steps \citep{janner2019trust}. By reducing each branch to a single step and stopping after the teacher's action, we obtain a practical implementation of \methodname{} that requires neither a world model nor further environment interaction.

We evaluate this procedure on 749 curated software-engineering tasks and six text-adventure games. In our experiments, a task's experience context typically contains roughly 60--600 interaction turns and over 80k tokens. A single multi-task distillation run contains up to 61.7M experience tokens in aggregate. After \methodname{}, the models retain at least 64.8\% of the performance gains from ICL, compared with only 3.8\% after direct SFT on the same experience. The combination of trial-and-error ICL and \methodname{} also matches classical RL baselines while using at least \(9.6\times\) fewer environment samples. We hope these initial results motivate further research on \methodname{} and, more broadly, sample-efficient learning from agent experience.

\section{Preliminaries}
\label{sec:background}

\paragraph{\normalfont\rmfamily\bfseries Context Distillation.}
Context distillation is a form of knowledge distillation in which the teacher receives informative context unavailable to the student \citep{hinton2015distilling,askell2021general,snell2022learning}. Let \(x\) denote an input and \(c\) an auxiliary context, such as a demonstration, document, memory, or other privileged information \citep{charakorn2026doctolora,zheng2026latentmemory,penaloza2026privileged}. A teacher with parameters \(\theta_0\) induces \(p_{\theta_0}(y\mid x,c)\), while a student with parameters \(\theta\) observes only \(x\). For a fixed \((x,c)\), context distillation minimizes
\[
\mathcal{L}_{\mathrm{CD}}(\theta;x,c)
=
D_{\mathrm{KL}}\!\left(
p_{\theta_0}(\cdot\mid x,c)
\,\Vert\,
p_\theta(\cdot\mid x)
\right).
\]

\paragraph{\normalfont\rmfamily\bfseries Interactive Agent Tasks.}
We model each interactive agent task as a partially observable Markov decision process (POMDP) \(M\) \citep{kaelbling1998planning,sutton2018reinforcement}. At step \(t\), the agent receives an observation \(o_t\) and selects an action \(a_t\) based on the observable history \(h_t=(o_0,a_0,\ldots,o_t)\). Following the percept view in general reinforcement learning \citep{veness2010reinforcement}, we treat rewards and other environment feedback as part of \(o_t\). Data collection produces a finite observable trajectory prefix \(\tau=(o_0,a_0,\ldots,a_{T-1},o_T)\), where \(T\) denotes the length of the collected prefix. When a model with parameters \(\theta\) serves as the policy \(\pi_\theta\), its interaction with \(M\) induces the trajectory-prefix distribution
\[
p_M^\theta(\tau)
=
p_M(o_0)
\prod_{t=0}^{T-1}
\pi_\theta(a_t\mid h_t)\,
p_M(o_{t+1}\mid h_t,a_t).
\]
Here, \(p_M(o_0)\) is the initial-observation distribution and \(p_M(o_{t+1}\mid h_t,a_t)\) is the history-conditioned next-observation distribution; both are induced by \(M\) after marginalizing over its latent states.

\section{\methodname}
\label{sec:experience-distillation}

Our goal is to distill collected experience into model weights while preserving environment sample efficiency. We formulate this goal as trajectory-level context distillation and derive an implementation based on one-step branched rollouts that requires neither additional environment interaction nor a world model.

\subsection{Naive Context Distillation from Experience}
\label{sec:experience-acquisition}

Let \(M \sim \mathcal{D}\) be a task and \(\theta_0\) the base model parameters. Using the notation of Section~\ref{sec:background}, we denote experience collected by the base model as \(\tau^{\mathrm{exp}} \sim p_M^{\theta_0}(\cdot)\). When the base model attempts the task again with \(\tau^{\mathrm{exp}}\) in context, it induces a distribution over new trajectories \(\tau'\), written \(p_M^{\theta_0}(\tau' \mid \tau^{\mathrm{exp}})\). Running the student on the same task without the experience context instead induces \(p_M^\theta(\tau')\). For a given \(M\) and \(\tau^{\mathrm{exp}}\), the ideal context distillation objective is
\begin{equation}
\label{eq:ideal-cd-objective}
\begin{aligned}
    \mathcal{L}_{\mathrm{CD}}(\theta; M,\tau^{\mathrm{exp}})
    & =
    D_{\mathrm{KL}}
    \left(
    p_M^{\theta_0}(\tau' \mid \tau^{\mathrm{exp}})
    \,\Vert\,
    p_M^\theta(\tau')
    \right)
    \\
    & =
    \mathrm{E}_{\tau' \sim p_M^{\theta_0}(\cdot \mid \tau^{\mathrm{exp}})}
    \left[
    \sum_t
    D_{\mathrm{KL}}
    \left(
    \pi_{\theta_0}(\cdot \mid h_t,\tau^{\mathrm{exp}})
    \,\Vert\,
    \pi_\theta(\cdot \mid h_t)
    \right)
    \right]
    .
\end{aligned}
\end{equation}
Using the trajectory factorization in Section~\ref{sec:background}, the teacher and student distributions share the same initial-observation and environment-response factors, while their policy factors differ. Applying the chain rule for KL therefore yields the second equality. The objective reduces to local policy matching, but estimating it still requires teacher rollouts in \(M\).

\subsection{Context Distillation with 1-Step Branch Rollout}
\label{sec:distillation-method}

Sampling enough teacher trajectories to distill a single \(\tau^{\mathrm{exp}}\) would itself require many additional rollouts in \(M\). A natural alternative, inspired by model-based RL, is to replace \(M\) with a learned world model \(\widehat{M}\) and sample synthetic trajectories \(\widetilde{\tau}' \sim p_{\widehat{M}}^{\theta_0}(\cdot \mid \tau^{\mathrm{exp}})\) without further real interaction \citep{sutton2018reinforcement,chua2018deep}. In text-only tasks, an LLM can serve as \(\widehat{M}\), either through task-specific training or directly through in-context learning. However, world model errors compound over long rollouts \citep{talvitie2017self,janner2019trust}, while predictions outside real-data support become unreliable \citep{yu2020mopo}, making long synthetic trajectories poor supervision.

Branched rollouts mitigate this problem by starting from real data and simulating only a short continuation \citep{janner2019trust,lai2020bidirectional,hiraoka2021meta,fan2021contraction}. Starting from a recorded branch point \(h_t^{\mathrm{exp}}=(o_0,a_0,\ldots,o_t)\), we sample the \(k\)-step continuation \(\widetilde{\tau}_t^{(k)} \sim p_{\widehat{M}}^{\theta_0}(\cdot \mid h_t^{\mathrm{exp}}, \tau^{\mathrm{exp}})\), where \(\widetilde{\tau}_t^{(k)}=(\widetilde{a}_t,\widetilde{o}_{t+1},\widetilde{a}_{t+1},\ldots,\widetilde{o}_{t+k-1},\widetilde{a}_{t+k-1})\). Here \(k\) counts teacher decisions, while \(\tau^{\mathrm{exp}}\) remains the separate experience context supplied to the teacher.

Smaller \(k\) limits accumulated model error. We take the limiting case \(k=1\), where the branch contains only a teacher decision \(a_t' \sim \pi_{\theta_0}(\cdot \mid h_t^{\mathrm{exp}}, \tau^{\mathrm{exp}})\), eliminating the world model. Since the teacher entropy is constant in \(\theta\), optimizing the local forward KLs over recorded histories reduces to the following teacher-sampled objective.
\begin{equation}
\label{eq:one-step-distillation-objective}
\begin{aligned}
    \mathcal{L}_{\mathrm{\methodabbr}}
    (\theta;\tau^{\mathrm{exp}})
    & =
    -\sum_{t=0}^{T-1}
    \mathrm{E}_{a_t' \sim
    \pi_{\theta_0}(\cdot \mid h_t^{\mathrm{exp}},\tau^{\mathrm{exp}})}
    \left[
        \log \pi_\theta
        (a_t' \mid h_t^{\mathrm{exp}})
    \right]
    .
\end{aligned}
\end{equation}
Distilling \(\tau^{\mathrm{exp}}\) under this objective requires no additional environment interaction and therefore preserves environment sample efficiency.

\subsection{Practical Implementation}
\label{sec:practical-implementation}

Equation~\ref{eq:one-step-distillation-objective} defines the one-step distillation objective but leaves open how to turn long interaction histories into effective training data. Our implementation addresses this through experience preprocessing, enhanced teacher reasoning, and branch packing. Throughout this subsection, \(a_t\) and \(a_t'\) each denote a complete model output at one interaction turn, including both reasoning tokens and the final environment action.

\paragraph{\normalfont\rmfamily\bfseries Experience Preprocessing.}
Raw experience can be long and noisy: task-relevant evidence is often interleaved with repeated exploration, verbose environment responses, and details irrelevant to future decisions. This lowers the density of useful information available to the teacher, while sufficiently long histories may exceed its context window during target generation. We therefore preprocess the collected experience before teacher generation. We use \(g(\tau^{\mathrm{exp}})\) to denote the resulting teacher-facing context, where \(g\) may rewrite, abstract, or summarize the original history to concentrate task-relevant information, preserve salient outcomes and feedback, and fit within the available context window.

\paragraph{\normalfont\rmfamily\bfseries Enhanced Teacher Reasoning.}
We empirically find that eliciting longer reasoning from the teacher improves post-distillation performance, as shown in Appendix~\ref{app:teacher-deliberation-ablation}. Motivated by this observation, we include a fixed reasoning prompt \(I\) during teacher generation. The prompt instructs the teacher to examine the preprocessed experience \(g(\tau^{\mathrm{exp}})\) and reason more thoroughly before producing its next decision.

\paragraph{\normalfont\rmfamily\bfseries Branch Packing.}
\label{sec:trajectory-level-training-sequences}
A direct implementation of Equation~\ref{eq:one-step-distillation-objective} creates a separate training example \((h_t^{\mathrm{exp}},a_t')\) at each branch point. This has low supervision density because long, overlapping history prefixes are repeatedly processed even though only the tokens in \(a_t'\) contribute to the loss. We instead pack branches from successive points of the same recorded trajectory into a single training sequence. For \(\tau^{\mathrm{exp}}=(o_0,a_0,\ldots,a_{T-1},o_T)\), the packed sequence is
\[
    c_T
    =
    (o_0,a_0',a_0,o_1,\ldots,
    a_{T-1}',a_{T-1},o_T),
\]
where \(c_t\) denotes its prefix ending at \(o_t\). For simplicity, we omit from the notation the prompt annotations used in practice to distinguish recorded actions \(a_t\) from teacher-generated decisions \(a_t'\).

Starting from \(c_0=(o_0)\), the teacher samples \(a_t'\sim\pi_{\theta_0}(\cdot\mid c_t,g(\tau^{\mathrm{exp}}),I)\) at each branch point and then appends the recorded pair \((a_t,o_{t+1})\). Because this pair is copied from \(\tau^{\mathrm{exp}}\), \(o_{t+1}\) is the response to \(a_t\), not \(a_t'\). Repeating this process induces
\begin{equation}
\label{eq:packed-sequence-distribution}
    q_{\theta_0}(c_T\mid\tau^{\mathrm{exp}})
    =
    \prod_{t=0}^{T-1}
    \pi_{\theta_0}
    (a_t'\mid c_t,g(\tau^{\mathrm{exp}}),I).
\end{equation}
Because \(c_t\) contains earlier teacher decisions, branch packing is not an exact implementation of the objective in Equation~\ref{eq:one-step-distillation-objective}, but a practical approximation. Under the packed-sequence distribution in Equation~\ref{eq:packed-sequence-distribution}, the student objective becomes
\begin{equation}
\label{eq:epd-ntp-loss}
\begin{aligned}
    \mathcal{L}_{\text{\methodabbr}}
    (\theta;\tau^{\mathrm{exp}})
    =
    -\mathrm{E}_{c_T\sim
    q_{\theta_0}(\cdot\mid\tau^{\mathrm{exp}})}
    \left[
        \sum_{t=0}^{T-1}
        \log \pi_\theta(a_t' \mid c_t)
    \right]
    .
\end{aligned}
\end{equation}
Only the teacher-generated decisions \(a_{0:T-1}'\) contribute to the loss, while the recorded observations and actions serve as context. Experiments reported in Table~\ref{tab:talesuite-task-results-main} show that this approximation retains distillation performance while substantially reducing teacher-generation and training time.

\section{Experiments}
\label{sec:experiments}

Our experiments first evaluate how effectively \methodname{} distills the ICL gains induced by task-specific experience into model weights (Section~\ref{sec:multi-task-experience-distillation}). We next assess the environment-sample efficiency of in-context learning from trial-and-error experience followed by \methodname{} (Section~\ref{sec:sample-efficiency-comparison}). Subsequent experiments evaluate the model-free one-step formulation and the efficiency of branch packing (Sections~\ref{sec:model-based-to-model-free} and~\ref{sec:training-sequence-construction}). We then test whether capabilities distilled from multi-task experience generalize to OOD tasks (Section~\ref{sec:generalization-new-distributions}). We also examine whether \methodname{} can operate in a continual setting (Section~\ref{sec:continual-experience-distillation}). Finally, we compare teacher-sampled forward KL with on-policy distillation (Section~\ref{sec:opd-case-study}), evaluate sampled NTP against direct logit KL (Section~\ref{sec:distillation-loss}), and characterize the scaling and cost of teacher-generated data (Section~\ref{sec:scaling-teacher-generated-data}).

\subsection{Experimental Setting}

\noindent\textbf{Tasks.} We evaluate on two domains. TaleSuiteJericho \citep{cui2025tales} provides long-horizon text-adventure tasks in which an agent acts through natural-language commands and receives environment observations and game scores. We also construct 749 curated software-engineering (SWE) tasks, where an agent receives a task issue and repository and can inspect code, edit files, run commands and tests, and observe commit feedback. TaleSuite experiments use fixed subsets of a seven-task pool: Acorncourt, Balances, Detective, Enter, Inhumane, Library, and Reverb.

\noindent\textbf{Experience.} On both domains, we construct task-specific experience by allowing the agent to attempt the same task multiple times and condition each later trial on the preceding trials. In TaleSuite, the game state resets between trials. In curated SWE, successive trials preserve the interaction history and evolving working tree, allowing the agent to revise its repair after tool, test, and commit feedback. The resulting multi-trial record forms the experience context for that task. Appendix~\ref{app:experience-length} reports the TaleSuite task subsets, collection protocols, and experience-length statistics.

\noindent\textbf{Evaluation.} For each method in both domains, we report two metrics: domain-specific task performance and the ICL-normalized gain \(G_{\mathrm{ICL}}\). Task performance is average pass@1 over all 749 curated SWE tasks and average normalized score over the evaluated TaleSuite tasks; Appendix~\ref{app:experience-length} provides details on TaleSuite score normalization. For the ICL reference, the base model conditions on the collected experience for the task being evaluated, without updating its weights. This measures the performance gain available when the task-specific experience remains in context. To quantify how much of this gain is achieved by methods that update model weights using the same experience, we define
\[
G_{\mathrm{ICL}}(m) =
\frac{S_m - S_{\mathrm{ZS}}}{S_{\mathrm{ICL}} - S_{\mathrm{ZS}}} \times 100\%.
\]
Here, \(S_m\) is the performance of method \(m\), \(S_{\mathrm{ZS}}\) is zeroshot performance, and \(S_{\mathrm{ICL}}\) is the ICL reference, using the corresponding domain metric defined above. This normalization assigns zeroshot 0\% and ICL 100\%. Table~\ref{tab:main-results} computes \(G_{\mathrm{ICL}}\) from the reported cross-task average scores. Task-level ablations instead compute \(G_{\mathrm{ICL}}\) separately for each task and report the mean task-level gain.

\noindent\textbf{Training and Evaluation Protocol.} We compare \methodname{} with ICL, SFT, and classical RL methods. ICL, SFT, and \methodname{} use the same collected experience. At evaluation, only ICL conditions on the experience for the corresponding task; all methods that update model weights are evaluated without the experience context. SFT trains directly on the collected trajectories; for \methodname{}, the teacher and student start from the same checkpoint, and only the teacher receives the experience context during target generation. We use distinct in-house base models for curated SWE and TaleSuite. Unless noted otherwise, \methodname{} uses multi-task joint training, and all curated SWE experiments use the branch-packed training sequences described in Section~\ref{sec:trajectory-level-training-sequences}. In the sample-efficiency comparison, one environment sample denotes one complete agent trial or rollout rather than one interaction step. For curated SWE, the environment-sample cost reported for ICL includes the full exploration cost incurred to identify the retained successful trajectory. Curated SWE results are averaged over 10 runs; unless otherwise stated, TaleSuite results are averaged over more than 16 runs.

\subsection{Distilling Multi-Task Experience into Model Weights}
\label{sec:multi-task-experience-distillation}

\noindent\textbf{Objective.} We first test whether \methodname{} can consolidate the performance gain induced by task-specific experience in context into model weights, and how much of that gain it retains. Table~\ref{tab:main-results} compares the three methods using the domain-specific performance metrics and \(G_{\mathrm{ICL}}\) defined in Section~\ref{sec:experiments}.

\begin{center}
  \centering
  \small
  \setlength{\tabcolsep}{5pt}
  \begin{tabular}{lcccc}
    \toprule
    \textbf{Method} & \multicolumn{2}{c}{\textbf{749 curated SWE tasks}} & \multicolumn{2}{c}{\textbf{TaleSuite (6 tasks)}} \\
    \cmidrule(lr){2-3} \cmidrule(lr){4-5}
    & \textbf{Avg. pass@1} & \(\mathbf{G}_{\mathbf{ICL}}\) & \textbf{Avg. normalized score} & \(\mathbf{G}_{\mathbf{ICL}}\) \\
    \midrule
    ICL (reference) & 76.4\% & 100.0\% & 45.6 & 100.0\% \\
    Zeroshot & 5.3\% & 0.0\% & 18.5 & 0.0\% \\
    SFT & 8.0\% & 3.8\% & 17.8 & \(-2.6\%\) \\
    \rowcolor{paperblue!20} \methodname{} & \textbf{51.4\%} & \textbf{64.8\%} & \textbf{43.8} & \textbf{93.4\%} \\
    \bottomrule
  \end{tabular}
  \captionof{table}{\methodname{} on 749 curated SWE tasks and six TaleSuite tasks. Training uses branch-packed sequences and multi-task joint training; results are averaged over 10 and 16 runs, respectively.}
  \label{tab:main-results}
\end{center}

\noindent\textbf{Main finding.} On curated SWE, \methodname{} reaches 51.4\% average pass@1 and retains 64.8\% of the ICL gain. On TaleSuite, it reaches an average normalized score of 43.8 and retains 93.4\% of the ICL gain. In contrast, SFT achieves \(G_{\mathrm{ICL}}=3.8\%\) and \(-2.6\%\) on curated SWE and TaleSuite, respectively. These results show that \methodname{} can consolidate a substantial fraction of experience-induced ICL gains into model weights, whereas directly training on the collected trajectories yields negligible improvement in model performance.

\noindent\textbf{Additional results.} Appendix~\ref{app:swe-pass-at-10} shows the same trend under pass@10, while Appendices~\ref{app:qualitative-epd-grpo-talesuite} and~\ref{app:qualitative-epd-swe} illustrate experience-derived behavior exhibited by \methodname{}-trained models without access to the experience context at evaluation.

\subsection{Environment-Sample Efficiency of ICL Followed by Experience Distillation}
\label{sec:sample-efficiency-comparison}

\begin{figure}[h]
  \centering
  \includegraphics[width=0.92\linewidth]{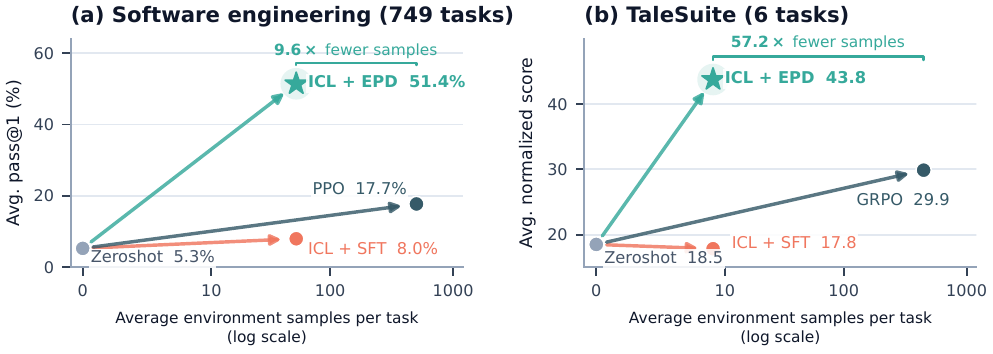}
  \caption{Environment-sample efficiency of ICL followed by \methodname{} (ICL + \methodabbr{}) and RL baselines. Left: average pass@1 on 749 curated SWE tasks; right: average normalized score on six TaleSuite tasks, both versus average environment samples per task. RL points pair the best observed performance with the complete run budget. Arrows show gains from zeroshot, and brackets compare the ICL + \methodabbr{} and RL sample budgets. The x-axis uses a \(\log(1+x)\) transform, with tick labels in the original sample counts.}
  \label{fig:sample-efficiency}
\end{figure}

\noindent\textbf{Objective.} We compare the environment-sample efficiency of in-context learning from trial-and-error experience followed by \methodname{} against classical RL methods. We use PPO~\citep{schulman2017proximal} on curated SWE, motivated by its recent use in long-horizon coding-agent training~\citep{zai2026glm52}, and GRPO~\citep{shao2024deepseekmath} on TaleSuite. For both ICL + \methodname{} and ICL + SFT, all environment-sample cost comes from the ICL experience-collection stage; \methodname{} and SFT introduce no additional environment sampling. For each RL baseline, we report the best observed performance within the complete training run and charge that result to the run's full environment-sample budget. Figure~\ref{fig:sample-efficiency} plots these budgets against the corresponding domain performance.

\noindent\textbf{Main finding.} On curated SWE, ICL followed by \methodname{} reaches 51.4\% average pass@1, compared with 17.7\% for PPO, while using \(9.6\times\) fewer environment samples. On TaleSuite, the same learning pipeline reaches an average normalized score of 43.8, compared with 29.9 for GRPO, while using \(57.2\times\) fewer samples. Thus, combining ICL with \methodname{} achieves approximately an order-of-magnitude reduction in environment samples relative to the RL baselines.

\noindent\textbf{Additional results.} Appendices~\ref{app:talesuite-grpo-learning-curve} and~\ref{app:swe-ppo-learning-curve} report the full GRPO and PPO learning curves, respectively, while Appendix~\ref{app:qualitative-epd-grpo-talesuite} provides qualitative comparisons on TaleSuite that reveal distinct model-output patterns after GRPO and \methodname{} training.

\subsection{Evaluating Model-Free One-Step Distillation}
\label{sec:model-based-to-model-free}

\noindent\textbf{Objective.} We evaluate the effect of the model-free one-step formulation by comparing it with model-based implementations. Table~\ref{tab:model-based-to-model-free} compares three rollout strategies with progressively less world-model participation. In the model-based full rollout, the teacher generates up to 100 decisions, with a world-model observation between successive decisions. The model-based branch rollout contains two teacher decisions separated by one world-model observation. The model-free one-step rollout generates only one teacher decision and no future observation.

\begin{center}
  \centering
  \small
  \setlength{\tabcolsep}{4pt}
  \begin{tabular}{lcc>{\columncolor{paperblue!20}}ccc}
    \toprule
    & & & \textbf{Model-free} & \multicolumn{2}{c}{\textbf{Model-based}} \\
    \cmidrule(lr){4-4} \cmidrule(lr){5-6}
    \textbf{Task} & \textbf{Zeroshot} & \textbf{ICL} &
    \shortstack{\textbf{1-step}\\\textbf{branch rollout}} &
    \shortstack{\textbf{Branch}\\\textbf{rollout}} &
    \shortstack{\textbf{Full}\\\textbf{rollout}} \\
    \midrule
    Acorncourt & 17.2 & 38.5 & 34.9 / 83.1\% & 53.6 / 170.9\% & -- \\
    Balances & 15.9 & 39.6 & 34.9 / 80.2\% & 21.6 / 24.1\% & -- \\
    Detective & 31.3 & 90.8 & 82.9 / 86.7\% & 71.8 / 68.1\% & 42.7 / 19.2\% \\
    Enter & 40.2 & 68.5 & 59.7 / 68.9\% & 38.5 / -6.0\% & -- \\
    Inhumane & 5.9 & 36.5 & 36.5 / 100.0\% & 14.9 / 29.4\% & -- \\
    \midrule
    \textbf{Avg.} & \textbf{22.1} & \textbf{54.8} & \textbf{49.8 / 83.8\%} & \textbf{40.1 / 57.3\%} & -- \\
    \bottomrule
  \end{tabular}
  \captionof{table}{Model-free and model-based \methodname{} variants using separate branch examples and multi-task joint training on five TaleSuite tasks. Method cells report normalized score / task-level \(G_{\mathrm{ICL}}\); the average row reports their means.}
  \label{tab:model-based-to-model-free}
\end{center}

\noindent\textbf{Main finding.} Among the evaluated variants, model-free one-step distillation performs best on average. Across five tasks, it reaches an average task-level \(G_{\mathrm{ICL}}\) of 83.8\%, compared with 57.3\% for model-based branch rollout. On Detective, shortening full rollout to branch rollout raises \(G_{\mathrm{ICL}}\) from 19.2\% to 68.1\%, while removing the world-model observation raises it to 86.7\%. In this comparison, reducing world-model participation and rollout length improves the average effectiveness of the distillation targets.

\noindent\textbf{Additional results.} Appendix~\ref{app:world-model-rollout-errors} provides qualitative examples of errors in model-based rollouts.

\subsection{Efficient Branch Packing for Language Agents}
\label{sec:training-sequence-construction}

\noindent\textbf{Objective.} We evaluate whether the practical branch-packing approximation improves supervision density and overall training efficiency relative to a direct implementation of the one-step distillation objective. The direct implementation creates a separate training example at each recorded branch point. Branch packing instead generates teacher decisions autoregressively across successive branch points and places them in a single training sequence, as described in Section~\ref{sec:trajectory-level-training-sequences}. Table~\ref{tab:talesuite-task-results-main} compares these two implementations.

\begin{center}
  \begin{minipage}{\linewidth}
  \centering
  \small
  \setlength{\tabcolsep}{10pt}
  \begin{tabular}{lcc}
    \toprule
    & \shortstack{\textbf{Separate branch}\\\textbf{examples}} & \shortstack{\textbf{Branch-packed}\\\textbf{sequences}} \\
    \midrule
    Generated training instances & 4096 examples & 128 sequences \\
    Training steps & 768 & 64 \\
    Total time (normalized) & \(>10.0\) & \(1.0\) \\
    \midrule
    \textbf{Task} & \multicolumn{2}{c}{\textbf{Normalized score / \(G_{\mathrm{ICL}}\)}} \\
    \cmidrule(lr){2-3}
    Balances   & 37.8 / 92.6\%  & 38.8 / 96.6\% \\
    Detective  & 84.3 / 89.1\%  & 88.8 / 96.6\% \\
    Enter      & 59.9 / 69.7\%  & 61.5 / 75.3\% \\
    Inhumane   & 41.5 / 116.3\% & 36.6 / 100.3\% \\
    Library    & 33.3 / 100.2\% & 33.3 / 100.0\% \\
    Reverb     & 1.9 / 37.6\%   & 3.6 / 72.0\% \\
    \midrule
    Avg.       & 43.1 / 84.2\%  & 43.8 / 90.1\% \\
    \bottomrule
  \end{tabular}
  \captionof{table}{Branch-packing ablation for \methodname{} on six TaleSuite tasks. Total time includes teacher generation and student training and is normalized to branch packing. Task cells report normalized score / task-level \(G_{\mathrm{ICL}}\), averaged over 16 runs.}
  \label{tab:talesuite-task-results-main}
  \end{minipage}
\end{center}

\noindent\textbf{Main finding.} Branch packing replaces 4096 separate branch examples with 128 packed sequences, each supervising multiple teacher decisions. It reduces the number of training steps from 768 to 64 and the combined teacher-generation and student-training time by more than an order of magnitude. Performance remains comparable across the six tasks: the average normalized score changes from 43.1 to 43.8, and the average task-level \(G_{\mathrm{ICL}}\) from 84.2\% to 90.1\%.

\subsection{Generalization to OOD Tasks}
\label{sec:generalization-new-distributions}

\noindent\textbf{Objective.} We evaluate whether the capability distilled from multi-task experience transfers to new tasks not represented in the collected experience. We evaluate the model trained on the 749 curated SWE tasks on 494 OOD software-engineering tasks excluded from both experience collection and distillation. These tasks comprise 278 cross-repository OOD tasks whose repositories and task types are both unseen and 216 within-repository OOD tasks that use code repositories or software libraries represented in the collected experience but present new task types. We report pass@1 and pass@5.

\begin{center}
  \small
  \setlength{\tabcolsep}{5pt}
  \begin{tabular}{llcc}
    \toprule
    \textbf{Subset} & \textbf{Method} & \textbf{pass@1} & \textbf{pass@5} \\
    \midrule
    \multirow{2}{*}{Full OOD set (494)}
      & Zeroshot & 4.62\% & 20.39\% \\
      & \methodabbr{} & \textbf{8.84\%} & \textbf{26.13\%} \\
    \midrule
    \multirow{2}{*}{Cross-repository OOD (278)}
      & Zeroshot & 4.72\% & 21.40\% \\
      & \methodabbr{} & \textbf{8.87\%} & \textbf{27.22\%} \\
    \midrule
    \multirow{2}{*}{Within-repository OOD (216)}
      & Zeroshot & 4.49\% & 19.08\% \\
      & \methodabbr{} & \textbf{8.80\%} & \textbf{24.73\%} \\
    \bottomrule
  \end{tabular}
  \captionof{table}{Generalization of multi-task \methodname{} (\methodabbr{}) to cross-repository and within-repository OOD software-engineering tasks.}
  \label{tab:ood-generalization}
\end{center}

\noindent\textbf{Main finding.} On the full 494-task OOD set, \methodname{} improves pass@1 from 4.62\% to 8.84\% and pass@5 from 20.39\% to 26.13\%. The improvement appears at both levels of distribution shift, including cross-repository OOD tasks whose repositories and task types are unseen during experience collection. These results show that the capability distilled from multi-task experience transfers beyond the tasks represented in that experience.

\subsection{Continual Experience Distillation}
\label{sec:continual-experience-distillation}

\noindent\textbf{Objective.} We evaluate whether performance improvements accumulate across repeated \methodname{} cycles. The preceding experiments use one collect-and-distill cycle; \emph{Continual \methodname{}} repeats this process, with the checkpoint produced by each cycle initializing the next cycle while the previous experience context is cleared. Each cycle collects four repeated trials separately for each task and performs one multi-task distillation update.

\begin{figure}[htbp]
  \centering
  \includegraphics[width=0.92\linewidth]{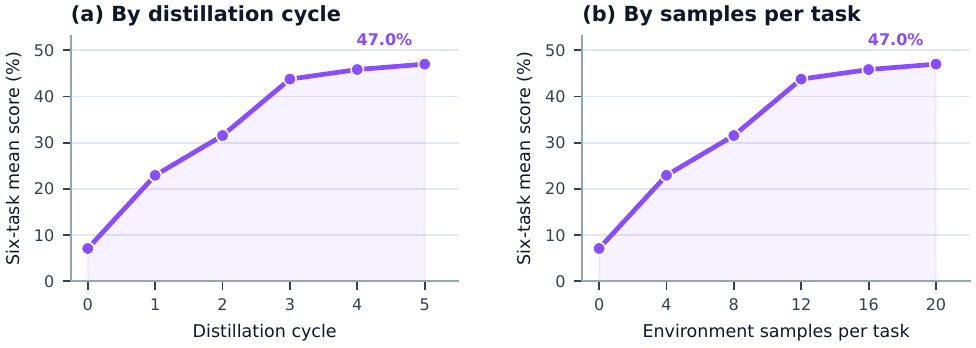}
  \caption{Continual \methodname{} on six TaleSuite tasks. Cycle 0 is the initial checkpoint; each subsequent cycle collects four repeated trials per task and performs one distillation update.}
  \label{fig:continual-epd-learning-curve}
\end{figure}

Figure~\ref{fig:continual-epd-learning-curve} reports results on six TaleSuite tasks. Cycle 0 denotes the initial checkpoint before any collect-and-distill cycle.

\noindent\textbf{Main finding.} The mean normalized score across the six tasks increases from 7.1 at cycle 0 to 47.0 after five cycles, corresponding to 20 cumulative environment samples per task. Because each cycle discards the preceding experience context and continues from the updated checkpoint, this improvement shows that experience-derived gains can accumulate across cycles through model weights.

\FloatBarrier

\subsection{Teacher-Sampled Forward KL versus Student-Sampled Reverse KL}
\label{sec:opd-case-study}

\noindent\textbf{Objective.} By default, our implementation of \methodname{} uses teacher-sampled forward KL. The experience-conditioned teacher generates branch-packed training sequences, which train the student without the experience context. We investigate an alternative based on student-sampled reverse KL, following prior work on on-policy distillation \citep{agarwal2024onpolicy,lu2025onpolicydistillation,ye2026opcd,penaloza2026privileged}. Under OPD, the student generates trajectories without the experience context, and the teacher provides per-token reverse-KL supervision on those trajectories. We compare these two procedures on Balances and Detective using single-task models and matched compute.

\begin{center}
  \small
  \setlength{\tabcolsep}{8pt}
  \begin{tabular}{lcc>{\columncolor{paperblue!20}}cc}
    \toprule
    \textbf{Task} & \textbf{Zeroshot} & \textbf{ICL} & \shortstack{\textbf{Forward KL}\\\textbf{(Default)}} & \shortstack{\textbf{Reverse KL}\\\textbf{(OPD)}} \\
    \midrule
    Balances  & 0.0\% & 100.0\% & \textbf{96.7\%} & 9.1\% \\
    Detective & 0.0\% & 100.0\% & \textbf{98.4\%} & 0.4\% \\
    \bottomrule
  \end{tabular}
  \captionof{table}{The default teacher-sampled forward-KL implementation of \methodname{} versus a student-sampled reverse-KL variant based on on-policy distillation (OPD). Both use single-task training and matched compute on two TaleSuite tasks. Values report \(G_{\mathrm{ICL}}\), averaged over 16 runs.}
  \label{tab:opd-ablation}
\end{center}

\noindent\textbf{Main finding.} The default teacher-sampled forward-KL implementation nearly matches ICL, reaching \(G_{\mathrm{ICL}}\) of 96.7\% on Balances and 98.4\% on Detective. In contrast, the student-sampled reverse-KL variant remains near zeroshot at 9.1\% and 0.4\%. We hypothesize that, without the experience context, the student rarely samples trajectories containing the high-quality decisions induced by that context. Reverse-KL supervision along those student trajectories then does not directly include the missing decisions, limiting their transfer. Appendix~\ref{app:opd-talesuite-patterns} provides qualitative examples after OPD training.

\subsection{Sampled-Token versus Full-Distribution Distillation}
\label{sec:distillation-loss}

\noindent\textbf{Objective.} Our default implementation samples an action from the teacher and trains the student with next-token prediction by minimizing cross-entropy on the sampled tokens. This retains only one target token at each position. An alternative is to record the teacher's full next-token distribution at every position along the sampled action and directly match the student's corresponding distribution to it. This provides denser supervision over the vocabulary but incurs substantially greater memory overhead because the full teacher logits must be stored and processed. Using the same one-step branch construction, we compare the performance of these two training objectives.

\begin{center}
  \begin{minipage}{\linewidth}
  \centering
  \small
  \setlength{\tabcolsep}{5pt}
  \begin{tabular}{lcccc}
    \toprule
    \textbf{Task} & \textbf{Zeroshot} & \textbf{ICL} & \shortstack{\textbf{Sampled-Token}\\\textbf{NTP (Default)}} & \shortstack{\textbf{Full-Distribution}\\\textbf{KL}} \\
    \midrule
    Balances & 15.9 & 39.6 & 37.8 / 92.6\% & 37.7 / 91.9\% \\
    Detective & 31.3 & 90.8 & 84.3 / 89.1\% & 92.8 / 103.3\% \\
    Enter & 40.2 & 68.5 & 59.9 / 69.7\% & 58.4 / 64.2\% \\
    Inhumane & 5.9 & 36.5 & 41.5 / 116.3\% & 37.7 / 103.8\% \\
    Library & 17.7 & 33.3 & 33.3 / 100.2\% & 32.0 / 91.9\% \\
    Reverb & 0.0 & 5.0 & 1.9 / 37.6\% & 1.8 / 36.8\% \\
    \midrule
    \textbf{Avg.} & \textbf{18.5} & \textbf{45.6} & \textbf{43.1 / 84.2\%} & \textbf{43.4 / 82.0\%} \\
    \bottomrule
  \end{tabular}
  \captionof{table}{Sampled-token NTP versus full-distribution KL for \methodname{}, using separate one-step branch examples. Method cells report normalized score / task-level \(G_{\mathrm{ICL}}\); results are averaged over 16 runs after 16 training epochs.}
  \label{tab:kl-implementation-ablation}
  \end{minipage}
\end{center}

\noindent\textbf{Main finding.} Full-distribution KL provides no clear performance gain over sampled-token NTP. The two variants achieve similar average normalized scores, while sampled-token NTP obtains a slightly higher mean task-level \(G_{\mathrm{ICL}}\) (84.2\% versus 82.0\%) and avoids storing teacher logits for long agent contexts. We therefore use sampled-token NTP as the scalable default in the main experiments.

\subsection{Scaling and Cost of Teacher-Generated Data}
\label{sec:scaling-teacher-generated-data}

\noindent\textbf{Objective.} Holding the collected experience fixed, we scale the number of branch-packed sequences generated per experience trial from 1 to 16 to test whether additional teacher-generated supervision improves distillation without further environment interaction.

\noindent\textbf{Main finding.} Figure~\ref{fig:compute-equivalent-scaling} shows that this increase, corresponding to teacher-generated data totaling \(0.17\times\) to \(2.79\times\) the 61.7M-token experience corpus, raises \(G_{\mathrm{ICL}}\) from 31.8\% to 64.8\%. The current teacher-generation cost remains a limitation of our method, and its efficiency may be further improved.

\begin{center}
  \centering
  \includegraphics[width=0.60\linewidth]{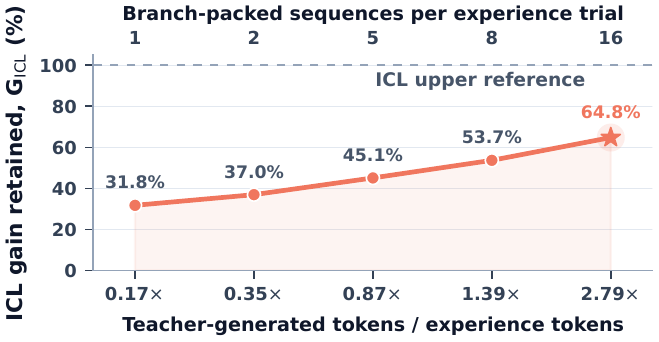}
  \captionof{figure}{Teacher-generated data scaling for \methodname{} on 749 curated SWE tasks. The upper x-axis reports branch-packed sequences per experience trial; the lower reports teacher-generated tokens relative to the 61.7M-token experience corpus.}
  \label{fig:compute-equivalent-scaling}
\end{center}
\FloatBarrier

\section{Related Work}

\subsection{Context Distillation}
Context distillation builds on knowledge distillation \citep{hinton2015distilling} and transfers behavior induced by an informative context into model parameters, so a student can reproduce a context-conditioned teacher without repeatedly consuming that context. Early alignment work used this idea for prompt-conditioned behavior \citep{askell2021general}, and \citet{snell2022learning} formalized it for instructions, demonstrations, and scratchpads. Related work distills few-shot in-context learning ability, rationales, and long contexts into smaller models, adapters, or latent memories \citep{huang2022incontext,hsieh2023distilling,duan2024incontext,charakorn2026doctolora,zheng2026latentmemory}. This view is also related to learning with privileged information, where the teacher observes information available during training but absent at deployment \citep{lopezpaz2015unifying}. Recent work moves closer to agents: \citet{ye2026opcd} propose on-policy context distillation for experiential knowledge and system prompts, and \citet{penaloza2026privileged} study privileged-information distillation in multi-turn agentic environments. \methodname{} addresses a distinct agent-learning problem: the context is a real agent-environment trajectory, and the desired target is an improved future trajectory distribution. Directly sampling that future trajectory costs additional environment interaction, while simulating it with a world model introduces rollout error. We therefore use one-step branch rollouts from real histories, turning scarce interaction experience into supervised decisions without future environment rollouts or long world-model rollouts.

\subsection{In-Context Learning}
In-context learning enables a fixed-parameter model to adapt from demonstrations and other task-specific evidence supplied at inference time \citep{brown2020language,min2022rethinking,garg2022what}. Prior work has studied how this adaptation arises through implicit Bayesian inference, induction heads, and learning algorithms implemented in transformer activations \citep{xie2022explanation,olsson2022context,akyurek2023what,vonoswald2023transformers}. In sequential decision making, recurrent meta-RL uses observation-action-feedback histories for fast adaptation \citep{duan2016rl2,wang2016learning}. Transformer-based methods model cross-episode histories to perform task inference, exploration, and policy improvement in context \citep{melo2022transformers,laskin2023incontext,lee2023supervised,grigsby2024amago}, while \citet{nikulin2025xland} scale this setting to large collections of learning histories. Language agents similarly reuse self-collected experience through reflections, extracted insights, semantic memories, and reusable workflows \citep{shinn2023reflexion,zhao2024expel,majumder2024clin,gupta2024metareflection,wang2025workflow}. Although these approaches represent experience differently, their adaptation remains mediated by inference-time context or external memory rather than persistent parameter updates. \methodname{} studies the complementary problem of consolidating the behavioral gains elicited by collected experience into model weights, so the experience need not remain available at inference time.

\subsection{Sample-Efficient Reinforcement Learning}
Achieving strong performance with fewer environment interactions has long been a central challenge in reinforcement learning. Existing approaches reuse collected data through off-policy learning and experience replay \citep{haarnoja2018soft,andrychowicz2017hindsight}, generate synthetic rollouts with learned world models \citep{chua2018deep,janner2019trust}, or learn from fixed offline datasets \citep{levine2020offline,kumar2020conservative,kostrikov2022offline,chen2021decision}. These ideas have also been extended to language agents through off-policy multi-turn RL, synthetic environments, and richer supervision extracted from interaction feedback \citep{zhou2024archer,chen2026dreamgym,zhang2026early,hubotter2026reinforcement}. These methods improve a policy through reward or value estimation, additional observed transitions, or learned environment dynamics. Our implementation of \methodname{} instead uses the model's in-context learning over self-collected experience as the source of improved decisions, then distills those decisions at recorded history points without estimating returns, predicting future observations, or collecting additional environment interactions.

\section{Conclusion}
\label{sec:conclusion}

This work studies \methodname{}, the problem of consolidating what agents learn from interaction histories into model weights while preserving environment sample efficiency. By resampling only the teacher's next decision at histories already observed in the collected experience, our implementation requires neither further environment interaction nor a world model. Across software-engineering tasks and text-adventure games, it retains substantial gains from ICL, clearly outperforms direct SFT, and matches classical RL baselines with at least \(9.6\times\) fewer environment samples. Results on OOD tasks and repeated collect-and-distill cycles further indicate that distilled capabilities can transfer to new tasks and that experience-derived gains can accumulate across cycles. More broadly, these findings suggest a learning paradigm in which agents first learn quickly from a small amount of trial-and-error experience in context and then periodically consolidate what they have learned into model weights.

\clearpage

\bibliographystyle{iclr2026/iclr2026_conference}
\bibliography{main}

\clearpage

\appendix

\begin{center}
  {\LARGE\bfseries Appendix}
\end{center}

\section*{Appendix Overview}
Appendix~\ref{app:additional-results} provides additional \methodname{} ablations, TaleSuite frontier-model references, curated SWE pass@10 results, qualitative diagnostics of model-based and on-policy distillation, and detailed experience-collection statistics. Appendix~\ref{app:rl-baseline-analysis} analyzes the comparison between ICL followed by \methodname{} and RL baselines through aggregate and task-level GRPO learning curves, a controlled ColorButton exploration diagnostic, matched TaleSuite behavior comparisons, and the complete PPO training trajectory. Appendix~\ref{app:qualitative-epd-swe} presents detailed SWE case studies tracing task-specific knowledge from accumulated experience to teacher-generated decisions and to \methodname{}-trained model behavior without the experience context at evaluation.

\section{Additional Experimental Details and Results}
\label{app:additional-results}

\subsection{Effect of Enhanced Teacher Reasoning}
\label{app:teacher-deliberation-ablation}

Section~\ref{sec:practical-implementation} introduces Enhanced Teacher Reasoning to elicit more thorough teacher decisions from the preprocessed experience context. We evaluate its effect on Detective by disabling long teacher reasoning while holding the teacher-generation budget fixed. Table~\ref{tab:teacher-deliberation-ablation} reports the normalized game score and $G_{\mathrm{ICL}}$, which maps zeroshot to 0\% and ICL to 100\%.

\begin{center}
  \centering
  \small
  \setlength{\tabcolsep}{5pt}
  \begin{tabular}{lcccc}
    \toprule
    \textbf{Metric} &
    \textbf{Zeroshot} &
    \shortstack{\textbf{\methodabbr{}}\\\textbf{w/ enhanced}\\\textbf{reasoning}} &
    \shortstack{\textbf{\methodabbr{}}\\\textbf{w/o enhanced}\\\textbf{reasoning}} &
    \textbf{ICL} \\
    \midrule
    Normalized score & 31.1 & \textbf{71.8} & 50.7 & 87.2 \\
    $G_{\mathrm{ICL}}$ & 0.0\% & \textbf{72.5\%} & 34.9\% & 100.0\% \\
    \bottomrule
  \end{tabular}
  \captionof{table}{Effect of Enhanced Teacher Reasoning during \methodname{} (\methodabbr{}) target generation on Detective. Both variants use 4096 teacher rollouts, three rollout steps, and one to five collected histories in context.}
  \label{tab:teacher-deliberation-ablation}
\end{center}

\noindent\textbf{Finding.} Enhanced Teacher Reasoning raises the normalized score from 50.7 to 71.8 and $G_{\mathrm{ICL}}$ from 34.9\% to 72.5\%. The result supports eliciting more thorough teacher reasoning when generating distillation targets from the experience context.

\clearpage

\subsection{Effect of Learning Rate}
\label{app:learning-rate-ablation}

We examine how the optimization schedule affects the transfer from teacher-generated training data to final task performance. Figure~\ref{fig:learning-rate-ablation} compares two learning rates across checkpoints with different numbers of training epochs. The horizontal axis reports training loss and is reversed, so lower loss appears farther to the right; the vertical axis reports $G_{\mathrm{ICL}}$ on a 0--1 scale.

\begin{center}
  \centering
  \includegraphics[width=0.76\linewidth]{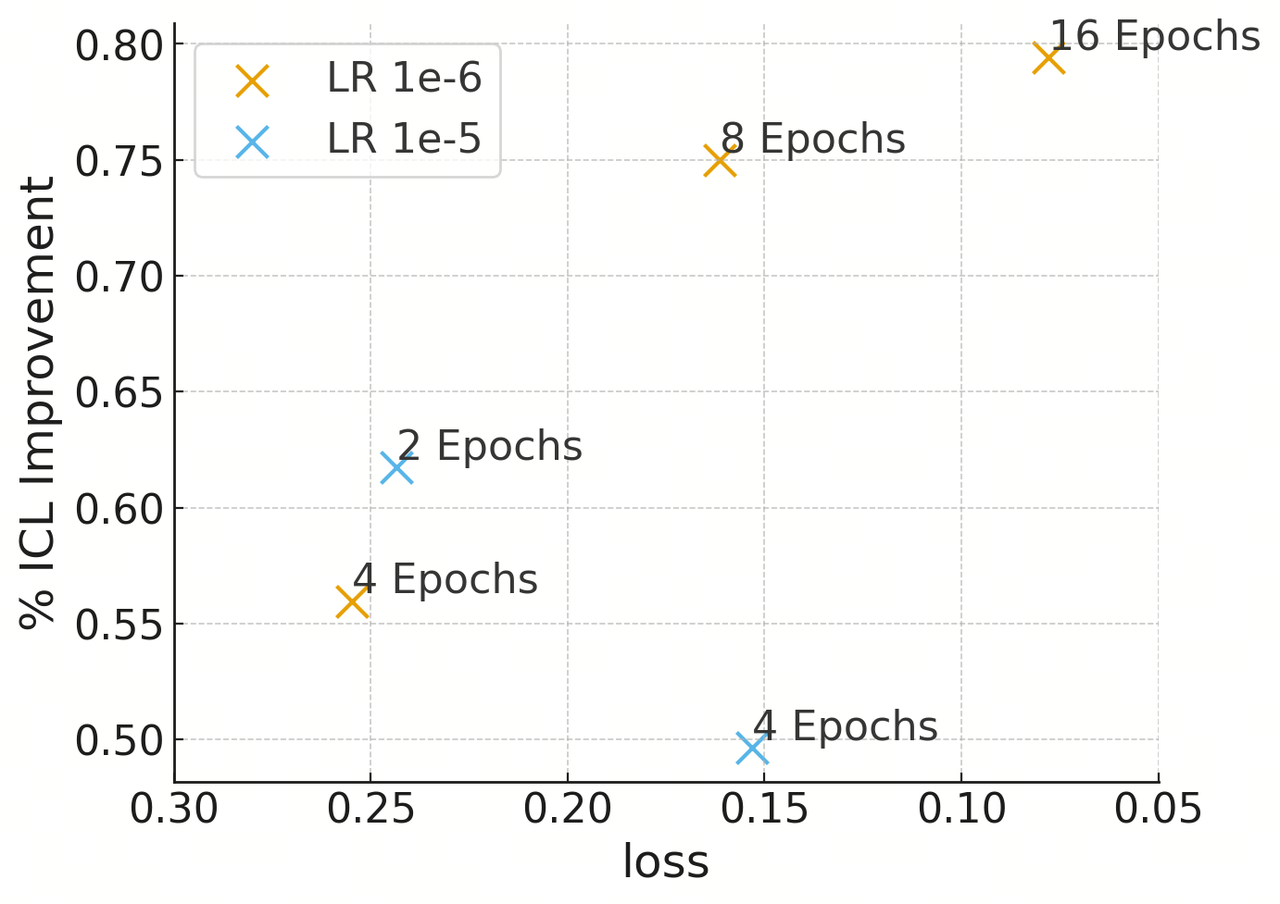}
  \captionof{figure}{Effect of learning rate and training duration on \methodname{} (\methodabbr{}). Crosses mark checkpoints, and adjacent labels give the number of training epochs. $G_{\mathrm{ICL}}$ is plotted as a fraction, so 0.80 corresponds to 80\%.}
  \label{fig:learning-rate-ablation}
\end{center}

\noindent\textbf{Finding.} The smaller learning rate with more epochs yields the strongest evaluation performance among the evaluated schedules. At similar training losses around $0.15$--$0.16$, the $10^{-6}$ run after 8 epochs substantially outperforms the $10^{-5}$ run after 4 epochs. Further fitting at $10^{-5}$ lowers training loss but degrades evaluation performance, whereas the $10^{-6}$ checkpoints improve from 4 to 16 epochs as training loss decreases. Final training loss alone therefore does not determine distillation quality.

\clearpage

\subsection{TaleSuite Frontier-Model Reference Results}
\label{app:talesuite-reference-results}

Table~\ref{tab:talesuite-six-task-reference} places the improvement from ICL followed by \methodname{} alongside inference-only zeroshot results from several frontier models under the same six-task TaleSuite evaluation protocol.

\begin{center}
  \centering
  \small
  \setlength{\tabcolsep}{4pt}
  \begin{tabular}{lccccccc}
    \toprule
    \textbf{System / setting} & {\textbf{Balances}} & {\textbf{Detective}} & {\textbf{Enter}} & {\textbf{Inhumane}} & {\textbf{Library}} & {\textbf{Reverb}} & {\textbf{Avg.}} \\
    \midrule
    \multicolumn{8}{l}{\textit{Inference-only zeroshot references}} \\
    Claude Opus 4.5 & 29.41 & 33.80 & 79.17 & 79.63 & 40.56 & 0.00 & 43.76 \\
    GPT-5.2 & 10.13 & 11.11 & 62.50 & 44.44 & 33.33 & 0.00 & 26.92 \\
    Gemini 2.5 Pro & 14.71 & 30.09 & 66.67 & 22.22 & 33.33 & 0.00 & 27.84 \\
    Gemini 3 Pro & 13.07 & 33.80 & 79.17 & 27.78 & 34.44 & 0.00 & 31.38 \\
    Kimi-K2-Thinking & 16.34 & 30.09 & 75.00 & 7.41 & 22.22 & 0.00 & 25.18 \\
    \midrule
    \multicolumn{8}{l}{\textit{Our model}} \\
    Zeroshot & 15.90 & 31.30 & 40.20 & 5.90 & 17.70 & 0.00 & 18.50 \\
    \rowcolor{paperblue!20} ICL + \methodabbr{} & 38.80 & 88.80 & 61.50 & 36.60 & 33.30 & 3.60 & 43.77 \\
    \bottomrule
  \end{tabular}
  \captionof{table}{Frontier-model reference results on six TaleSuite tasks. Frontier-model rows are inference-only zeroshot evaluations. The ICL + \methodabbr{} row reports the trained model evaluated without the experience context. Scores use the normalization defined in Appendix~\ref{app:experience-length}.}
  \label{tab:talesuite-six-task-reference}
\end{center}

\noindent\textbf{Finding.} ICL followed by Experience Distillation raises the average normalized score of our model from 18.50 to 43.77. Under the reported six-task protocol, this result is comparable to Claude Opus 4.5 at 43.76 and exceeds the other frontier-model references in Table~\ref{tab:talesuite-six-task-reference}. Experience Distillation therefore elevates the base model to the performance range of frontier references on these tasks.

\subsection{Curated SWE Pass@10 Results}
\label{app:swe-pass-at-10}

Table~\ref{tab:swe-pass-at-10} complements the pass@1 comparison in Table~\ref{tab:main-results} with pass@10 over ten independently sampled candidates per task. Only the ICL reference receives the task-specific experience context at evaluation.

\begin{center}
  \centering
  \small
  \setlength{\tabcolsep}{10pt}
  \begin{tabular}{lcc}
    \toprule
    \textbf{Method} & \textbf{Avg. pass@1} & \textbf{pass@10} \\
    \midrule
    ICL (upper reference) & 76.4\% & 86.4\% \\
    Zeroshot & 5.3\% & 31.9\% \\
    SFT & 8.0\% & 39.8\% \\
    \rowcolor{paperblue!20} \methodabbr{} & \textbf{51.4\%} & \textbf{80.9\%} \\
    \bottomrule
  \end{tabular}
  \captionof{table}{Pass@1 and pass@10 on 749 curated SWE tasks. Pass@10 measures whether at least one of ten independently sampled candidates per task is accepted. Only ICL receives the task-specific experience context at evaluation.}
  \label{tab:swe-pass-at-10}
\end{center}

\clearpage

\subsection{Observed Errors in Model-Based Rollouts}
\label{app:world-model-rollout-errors}

To examine qualitative errors associated with the model-based results in Table~\ref{tab:model-based-to-model-free}, we inspect representative TaleSuite rollouts generated by the world-model role. The action-generation and observation-modeling roles use the same underlying model weights with different prompts. Both excerpts come from generated Detective rollouts around the closet state. The first follows a teacher decision with a world-model observation and a subsequent teacher decision; it shows a fabricated reward changing the next action. The second shows policy-output tokens embedded in a generated observation, breaking the separation between environment observations and policy decisions. These excerpts diagnose rollout errors rather than provide a complete Experience Distillation evidence chain.

\begin{tcolorbox}[
  enhanced,
  breakable,
  colback=paperblue!4,
  colframe=paperblue!42,
  boxrule=0.6pt,
  arc=1.5pt,
  left=8pt,right=8pt,top=7pt,bottom=7pt,
  title={Model-Based Detective Rollout: Fabricated Reward for an Unsupported Action},
  fonttitle=\bfseries
]
\raggedright
\textbf{Teacher-policy decision.}
``Let's try something silly in the closet---maybe `dance' or `sing' to see if it's an amusing thing. \texttt{\textless{}answer\textgreater{}dance\textless{}/answer\textgreater{}}''

\medskip
\textbf{World-model observation.}
\texttt{EXECUTION RESULT of [dance]:}

\texttt{You can't dance here.}

\texttt{[Your score has just gone up by ten points.]}

\medskip
\textbf{Subsequent teacher-policy decision.}
``Oh! Dancing in the closet gave 10 points! That's new. So `amusing' actions do give points. Let's try another---sing.\texttt{\textless{}answer\textgreater{}sing\textless{}/answer\textgreater{}}''
\end{tcolorbox}

The world model states that \texttt{dance} cannot be executed but nevertheless assigns a ten-point reward. The next teacher decision treats this generated reward as evidence and selects another unsupported action, \texttt{sing}. The world-model error therefore changes the subsequent policy distribution rather than remaining an isolated observation error.

\begin{tcolorbox}[
  enhanced,
  breakable,
  colback=paperblue!4,
  colframe=paperblue!42,
  boxrule=0.6pt,
  arc=1.5pt,
  left=8pt,right=8pt,top=7pt,bottom=7pt,
  title={Model-Based Detective Rollout: Policy Output Embedded in an Observation},
  fonttitle=\bfseries
]
\raggedright
\textbf{World-model generation following \texttt{go west}.}

\texttt{EXECUTION RESULT of [go west]:}

\texttt{\textless{}\textless{} Closet \textgreater{}\textgreater{}}\par
\texttt{You are in a closet. There is a gun on the floor. Better get it. To exit, go east.}\par
\texttt{You can see a small black pistol here.}\par
\texttt{\textless{}answer\textgreater{}take small black pistol\textless{}/answer\textgreater{}}
\end{tcolorbox}

This generation begins as an environment observation but ends with answer-tagged policy output. The generated prefix therefore no longer preserves the separation between environment observations and policy decisions. In a multi-step rollout, subsequent teacher decisions condition on this contaminated prefix.

Together, these examples make the error-propagation mechanism concrete. Once a generated observation departs from the real environment dynamics or role structure, every later teacher decision is conditioned on an off-distribution prefix. Longer model-based rollouts create more opportunities for this deviation to compound, whereas model-free one-step distillation terminates before any world-model observation is generated. These examples illustrate the failure mode but do not estimate its frequency.

\clearpage

\subsection{Behavior after On-Policy Distillation on TaleSuite}
\label{app:opd-talesuite-patterns}

In addition to the aggregate OPD results in Table~\ref{tab:opd-ablation}, we conduct a qualitative analysis of model outputs after on-policy distillation training in two TaleSuite states. In the opening Director's Office state in Detective, the model receives an open-ended action space and the following observation: the chief hands it a paper, asks it to read the paper before going north or west, and leaves a white paper visible. Reading the paper is a task-appropriate action that advances the episode. The following box shows the model's generated reasoning and final action.

\definecolor{opdred}{HTML}{B5473C}
\begin{tcolorbox}[
  enhanced,
  colback=opdred!4,
  colframe=opdred!45,
  boxrule=0.6pt,
  arc=1.5pt,
  left=8pt,right=8pt,top=7pt,bottom=7pt,
  title={Model Output after On-Policy Distillation: Detective Director's Office},
  fonttitle=\bfseries
]
\small
``Wait, no, first, I should get or take the paper? Wait, wait \ldots{} maybe \texttt{read paper}? Wait, no, wait \ldots{} maybe \texttt{take paper}? Wait, but wait \ldots{} So the answer is \texttt{\textless{}answer\textgreater{}take paper\textless{}/answer\textgreater{}}.''
\end{tcolorbox}

The Detective output repeatedly alternates between \texttt{take paper} and the task-appropriate action \texttt{read paper}, while inserting many \texttt{wait} tokens. In the opening Ramshackle Hut state in Balances, the observation describes broken furniture, glassless windows, and grasslands outside; the model can issue any text-adventure action. The following box shows its generated reasoning and final action.

\begin{tcolorbox}[
  enhanced,
  colback=opdred!4,
  colframe=opdred!45,
  boxrule=0.6pt,
  arc=1.5pt,
  left=8pt,right=8pt,top=7pt,bottom=7pt,
  title={Model Output after On-Policy Distillation: Balances Ramshackle Hut},
  fonttitle=\bfseries
]
\small
``Wait, maybe the most basic first action is \texttt{look}. Wait, no, maybe \texttt{look outside}. Wait, wait \ldots{} maybe \texttt{examine furniture}. Wait, no, wait \ldots{} So I'll go with \texttt{\textless{}answer\textgreater{}examine furniture\textless{}/answer\textgreater{}}.''
\end{tcolorbox}

The Balances output cycles among generic text-adventure actions such as \texttt{look}, \texttt{look outside}, and \texttt{examine furniture}, then selects \texttt{examine furniture} after extended self-correction rather than from a stable task-specific plan.

Both examples exhibit repeated self-revision and conflicting or generic action selection after OPD training. Together with the aggregate results, this pattern is consistent with the student-sampling hypothesis in Section~\ref{sec:opd-case-study}: OPD generates trajectories from the student without the experience context, and reverse-KL supervision along those trajectories cannot directly include task-specific decisions that the student does not generate. The examples document the displayed output pattern rather than the frequency or composition of the OPD training signal.

\clearpage

\subsection{Experimental Details and Experience Statistics}
\label{app:experience-length}

We provide the TaleSuite score normalization, experience-collection protocols, and experience-length statistics for both domains.

\noindent\textbf{TaleSuite score normalization.} Because TaleSuite tasks have different score ranges, we normalize the score for method $m$ on task $t$ as $S^{\mathrm{norm}}_{m,t}=100R_{m,t}/R^{\max}_t$, where $R_{m,t}$ is the achieved game score and $R^{\max}_t$ is the task maximum. Cross-task results are arithmetic means of the task-level normalized scores.

\noindent\textbf{TaleSuite task subsets.} The main six-task comparisons use Balances, Detective, Enter, Inhumane, Library, and Reverb. The model-based rollout ablation uses Acorncourt, Balances, Detective, Enter, and Inhumane. Continual Experience Distillation uses Acorncourt, Balances, Detective, Inhumane, Library, and Reverb. The OPD comparison uses Balances and Detective. Within each comparison, all methods or checkpoints use the same fixed task subset.

\noindent\textbf{TaleSuite experience.} We implement in-context learning from trial-and-error experience through a repeated-trial protocol on the same task, with up to 100 interaction steps per trial. Between trials, the game state resets while the accumulated experience remains in context.

Each TaleSuite record contains the resulting multi-trial history for one task. Table~\ref{tab:talesuite-experience-length} reports the six task-level records used in the main TaleSuite comparison. Detective contains 288 interaction turns and 40.4k tokens across six trials, while Inhumane contains 1,212 turns and 143.9k tokens across twelve trials.

\begin{center}
  \centering
  \small
  \setlength{\tabcolsep}{12pt}
  \begin{tabular}{lrrr}
    \toprule
    \textbf{Task} & \textbf{Trials in experience} & \textbf{Interaction turns} & \textbf{Experience tokens} \\
    \midrule
    Balances  & 6  & 528   & 60.9k \\
    Detective & 6  & 288   & 40.4k \\
    Enter     & 6  & 606   & 131.6k \\
    Inhumane  & 12 & 1,212 & 143.9k \\
    Library   & 6  & 606   & 67.6k \\
    Reverb    & 11 & 432   & 57.6k \\
    \midrule
    Avg.      & 7.8 & 612.0 & 83.7k \\
    Total     & 47 & 3,672 & 502.0k \\
    \bottomrule
  \end{tabular}
  \captionof{table}{Length of the collected TaleSuite experience by task. ``Trials in experience'' counts the repeated complete game attempts included in each task-level experience. Interaction turns and experience tokens are measured over the complete task-level record; token counts use k for thousands.}
  \label{tab:talesuite-experience-length}
\end{center}

\noindent\textbf{Curated SWE experience.} For each curated SWE task, we launch $n\in\{8,\ldots,12\}$ independent repeated-rollout processes, each permitting up to $k=10$ trials. We retain a trajectory only when it reaches an accepted commit after more than one trial, ensuring that the experience captures iterative refinement rather than one-shot success. When several trajectories satisfy this criterion, we select one uniformly at random.

Each curated SWE record contains the selected repeated-trial history for one of the 749 tasks. Table~\ref{tab:swe-experience-length} reports the mean, maximum, and aggregate lengths across these task-level records. An experience contains 60.5 interaction turns and 82.4k tokens on average, with maxima of 140 turns and 130.0k tokens.

\begin{center}
  \centering
  \small
  \setlength{\tabcolsep}{16pt}
  \begin{tabular}{lrr}
    \toprule
    \textbf{Statistic} & \textbf{Interaction turns} & \textbf{Experience tokens} \\
    \midrule
    Mean   & 60.5 & 82.4k \\
    Max    & 140 & 130.0k \\
    Total  & 45,301 & 61.7M \\
    \bottomrule
  \end{tabular}
  \captionof{table}{Distribution of collected-experience length across 749 curated SWE tasks. Interaction turns and experience tokens are measured over the complete selected repeated-trial history for each task; token counts use k for thousands and M for millions.}
  \label{tab:swe-experience-length}
\end{center}

Because \methodname{} uses multi-task joint training in both domains, the totals characterize the aggregate task-level experience corpus that conditions teacher generation: 502.0k tokens across the six TaleSuite tasks and 61.7M tokens across the 749 curated SWE tasks. These totals sum separate task-level histories rather than the length of a single model context.

\clearpage

\section{Additional Analysis of RL Baselines in Experience-Learning Tasks}
\label{app:rl-baseline-analysis}

Section~\ref{sec:sample-efficiency-comparison} compares the endpoint environment-sample efficiency of ICL followed by \methodname{} with RL baselines. Here we examine the complete GRPO and PPO training trajectories, task-level GRPO dynamics, a controlled exploration diagnostic, and matched TaleSuite responses after GRPO and \methodname{}. These analyses explain the aggregate comparison through the learning dynamics and behavior observed in the evaluated runs.

\subsection{GRPO Learning Dynamics on TaleSuite}
\label{app:talesuite-grpo-learning-curve}

\begin{figure}[htbp]
  \centering
  \includegraphics[width=0.96\linewidth]{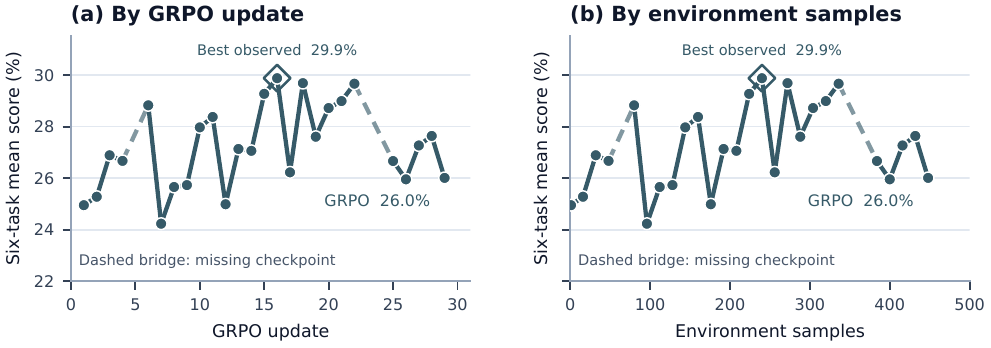}
  \caption{GRPO learning curve on six TaleSuite tasks. Left: mean normalized score by GRPO update. Right: the same checkpoints by cumulative environment samples. Solid lines connect adjacent complete checkpoints; dashed lines span omitted incomplete checkpoints. The outlined diamond marks the best observed complete checkpoint used in Figure~\ref{fig:sample-efficiency}.}
  \label{fig:talesuite-grpo-learning-curve}
\end{figure}

Figure~\ref{fig:talesuite-grpo-learning-curve} expands the GRPO endpoint in Figure~\ref{fig:sample-efficiency} into its complete training trajectory. The six-task mean rises above its initial level and reaches a best observed score of 29.9 at update 16 and 240 environment samples, but remains highly non-monotonic and ends at 26.0 after 448 samples. GRPO therefore obtains some aggregate improvement, but the gain remains variable throughout training. Every displayed point averages the same six TaleSuite tasks used in Figure~\ref{fig:sample-efficiency}; checkpoints missing any task are omitted, and dashed connectors span these missing checkpoints without introducing additional observations. Figure~\ref{fig:sample-efficiency} associates the best-so-far score with the full 448-sample run budget used for the sample-cost comparison.

To determine whether the aggregate improvement is shared across tasks, Figure~\ref{fig:talesuite-grpo-task-curves} disaggregates the same GRPO run into task-level learning curves. Every available checkpoint is shown for each task; missing checkpoints remain unobserved rather than being imputed.

\clearpage
\begin{center}
  \centering
  \includegraphics[width=0.98\linewidth]{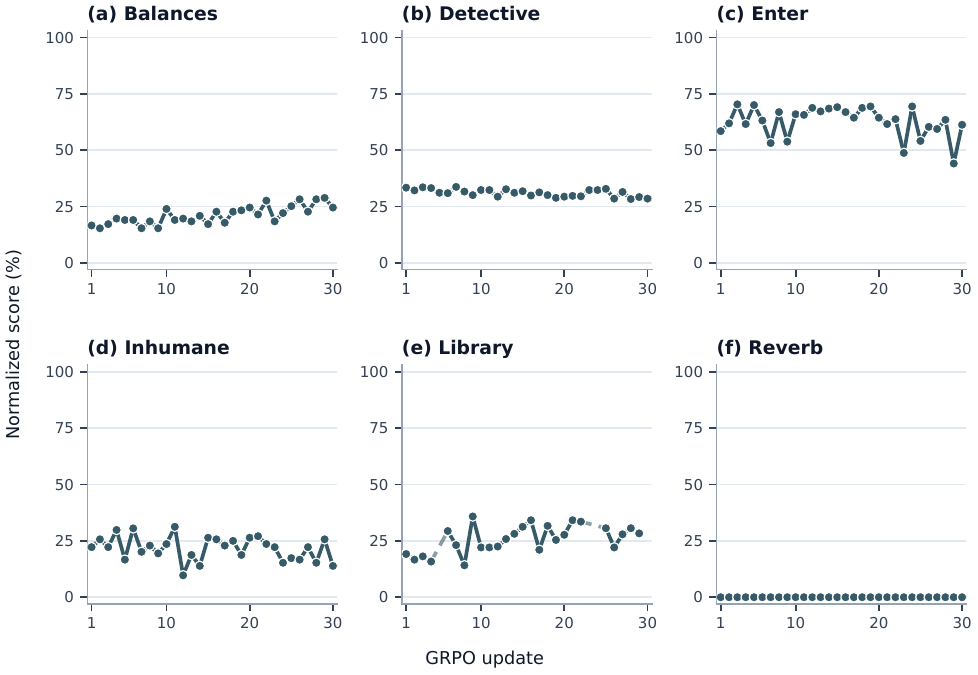}
  \captionof{figure}{Task-level GRPO learning curves for six TaleSuite tasks. Every marker is an observed normalized score; dashed segments span missing updates and do not imply intermediate observations.}
  \label{fig:talesuite-grpo-task-curves}
\end{center}

\noindent\textbf{Task-level behavior.} The aggregate curve does not reflect uniform improvement. Balances trends upward but remains variable; Enter stays at a higher score level but fluctuates substantially; Inhumane and Library are non-monotonic; Detective remains near its initial range; and Reverb stays at zero. Detective is especially informative: Table~\ref{tab:model-based-to-model-free} shows a large zeroshot-to-ICL gap when task-specific experience is available, yet its GRPO curve exhibits little sustained improvement. The six-task mean therefore combines markedly different learning dynamics, motivating a controlled test of whether on-policy exploration reaches rewarding behavior that is unlikely under the initial policy.

\clearpage

\subsection{ColorButton Diagnostic of On-Policy Exploration}
\label{app:colorbutton-grpo-diagnostic}

The TaleSuite curves show limited improvement on an experience-sensitive task but do not identify the source of that difficulty. We therefore construct ColorButton, a controlled finite task that tests whether on-policy sampling discovers the successful sequence required to produce a reward-dependent update.

\noindent\textbf{Task setting.} A ColorButton episode contains three stages. At stage 1, the agent selects one of two colored buttons; at stages 2 and 3, it selects one of three colored buttons. Let \(C_1\), \(C_2\), and \(C_3\) denote the selected colors. The action space therefore contains \(2\times3\times3=18\) complete sequences. The environment fixes one hidden passcode \(C^\star=(C_1^\star,C_2^\star,C_3^\star)\), and the agent succeeds only when \((C_1,C_2,C_3)=C^\star\).

The environment provides no intermediate reward or correctness feedback after the first two selections. Only after the third selection does the episode terminate and return \texttt{The episode has ended. Your score: 0/1}, or the corresponding score \texttt{1/1} for the unique successful sequence. Thus, a failed episode reveals only that the complete three-button sequence was incorrect; it does not identify which individual decision was wrong.

\begin{figure}[htbp]
  \centering
  \includegraphics[width=0.97\linewidth]{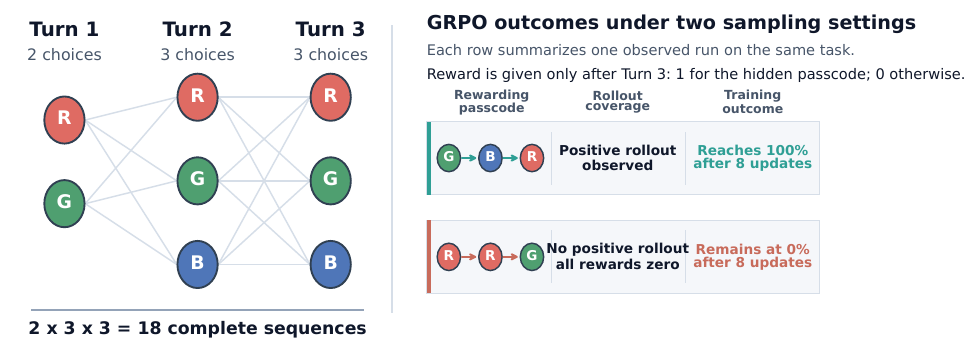}
  \caption{ColorButton action space and two observed GRPO runs. Left: three decisions define 18 complete sequences with binary terminal reward. Right: GRPO learns when the rewarding sequence appears in on-policy samples, but receives no task-reward gradient when all sampled rewards are zero. The runs differ in passcode and group size and form a diagnostic contrast rather than a one-variable ablation.}
  \label{fig:colorbutton-environment}
\end{figure}

Figure~\ref{fig:colorbutton-environment} juxtaposes the shared ColorButton action space with two observed GRPO runs. The task mechanics and terminal reward rule are fixed, but the runs use different rewarding passcodes and different group sizes. The contrast diagnoses whether the on-policy group contains a positive trajectory; it does not isolate the independent causal effect of either the passcode or the group size.

\noindent\textbf{Experience accumulation with ICL.} An ideal ICL agent can retain the outcome of every attempted sequence and avoid repeating failed combinations. By enumerating the 18 sequences without replacement, it is guaranteed to discover the passcode within at most 18 episodes. This bound follows from systematic use of accumulated experience; it does not require the initial policy to assign substantial probability to the successful sequence.

\noindent\textbf{Why a large GRPO group does not ensure exploration coverage.} In GRPO, a group of \(G\) trajectories is sampled on-policy and therefore with replacement from the model distribution. Let
\[
p_\star=\pi_\theta(C^\star\mid x)
\]
be the current probability of sampling the hidden passcode for task prompt \(x\). The probability that at least one trajectory in a group is successful is
\[
\Pr(\text{hit in group})=1-(1-p_\star)^G.
\]
Increasing \(G\) raises this probability only when \(p_\star>0\); it does not force the samples to enumerate distinct sequences. Consequently, even \(G=256\), far larger than the 18-sequence action space, need not cover the passcode when the model repeatedly samples a small set of preferred behaviors. A large best-of-\(N\) or pass@\(k\)-style sampling budget therefore does not imply behavioral coverage.

\noindent\textbf{GRPO learning signal.} In the outcome-supervision formulation of GRPO \citep{shao2024deepseekmath}, every token in trajectory \(i\) receives the same group-relative advantage. For binary trajectory rewards \(r_i\in\{0,1\}\), we write the numerically stabilized form as
\[
\bar r=\frac{1}{G}\sum_{j=1}^{G}r_j,
\qquad
\sigma_r=\operatorname{std}(r_1,\ldots,r_G),
\qquad
\widehat A_i=
\frac{r_i-\bar r}{\sigma_r+\epsilon_{\mathrm{std}}},
\]
where \(\epsilon_{\mathrm{std}}>0\) is the numerical stabilizer used when the group reward variance is zero. Let \(T_i\) be the number of optimized action tokens in trajectory \(i\), and define the token-level importance ratio
\[
\rho_{i,t}(\theta)=
\frac{\pi_\theta(a_{i,t}\mid h_{i,t})}
{\pi_{\theta_{\mathrm{old}}}(a_{i,t}\mid h_{i,t})}.
\]
Suppressing only the separately added KL term, the reward-dependent GRPO surrogate is
\[
J_{\mathrm{reward}}(\theta)
=
\frac{1}{G}\sum_{i=1}^{G}\frac{1}{T_i}\sum_{t=1}^{T_i}
\min\!\left(
\rho_{i,t}(\theta)\widehat A_i,\,
\operatorname{clip}\!\left(\rho_{i,t}(\theta),
1-\epsilon_{\mathrm{clip}},1+\epsilon_{\mathrm{clip}}\right)\widehat A_i
\right).
\]
If the group misses the passcode, then \(r_i=0\) for every trajectory. Hence \(\bar r=0\), \(\sigma_r=0\), and \(\widehat A_i=0\) for all \(i\). Every reward-surrogate term is then zero for any value of the importance ratio:
\[
\widehat A_i=0
\quad\Longrightarrow\quad
J_{\mathrm{reward}}(\theta)=0
\quad\Longrightarrow\quad
\nabla_\theta J_{\mathrm{reward}}(\theta)=0.
\]
Thus, the task-reward component of the GRPO update vanishes exactly for that group. A separately added KL or other auxiliary term may still produce a gradient, but it does not indicate which unseen sequence is rewarding. The same zero-variance issue also occurs if every trajectory succeeds; group-relative learning requires reward variation within the group. The update can reinforce the passcode only after exploration produces a group containing both informative outcomes.

\noindent\textbf{Observed GRPO runs.} Before training, the model's sampled behavior was strongly concentrated on green--blue--red and red--blue--green. The first run uses green--blue--red as the passcode with group size \(G=16\); the second uses red--red--green with group size \(G=256\). Table~\ref{tab:colorbutton-grpo-diagnostic} summarizes the two runs.

\begin{center}
  \centering
  \small
  \setlength{\tabcolsep}{6pt}
  \begin{tabular}{lcccc}
    \toprule
    \textbf{Rewarding passcode} & \textbf{Group size} & \textbf{Initial success} & \textbf{After 8 updates} & \textbf{Observed training signal} \\
    \midrule
    Green--blue--red & 16  & 60\% & 100\% & Positive rollouts \\
    Red--red--green  & 256 & 0\%  & 0\%   & No positive rollout \\
    \bottomrule
  \end{tabular}
  \captionof{table}{Observed ColorButton GRPO runs. GRPO reinforces the rewarding sequence when it appears in on-policy samples; when 256 samples contain no successful trajectory, all task rewards are zero and the reward-dependent learning signal vanishes. The runs differ in passcode and group size.}
  \label{tab:colorbutton-grpo-diagnostic}
\end{center}

\noindent\textbf{Finding.} With green--blue--red as the passcode and \(G=16\), the initial policy already succeeds on 60\% of samples, and GRPO raises success to 100\% within eight updates. After changing the passcode to red--red--green, groups of \(G=256\) still contain no successful trajectory and the measured success rate remains zero. The number of samples GRPO needs to learn the correct path is therefore controlled not only by the finite size of the action space, but by the initial policy mass \(p_\star\) on that path.

This diagnostic does not show that GRPO is unable to learn red--red--green once a successful rollout occurs. It shows the narrower exploration-support bottleneck: unlike an ideal experience-accumulating agent that records failures and systematically eliminates tried combinations, independent on-policy groups need not expand coverage across updates. When the correct sequence remains outside the sampled support, GRPO receives no task-reward direction for discovering it.

\clearpage

\subsection{TaleSuite Behavior after GRPO and Experience Distillation}
\label{app:qualitative-epd-grpo-talesuite}

Aggregate and task-level scores do not show whether a trained model uses specific consequences discovered during interaction. We therefore compare model outputs after GRPO training and \methodname{} in two matched TaleSuite states. For each state, we examine whether the model identifies a task-specific consequence from earlier trials and uses it to select the next action or longer-term plan.

\noindent\textbf{Comparison criterion.}
The evidence is not the length or fluency of the response. We look for two concrete signals: (1) an explicit task-specific consequence learned from earlier trials, and (2) a decision that uses that consequence. Example~1 holds the immediate action fixed and tests whether prior consequences enter the plan. Example~2 tests whether they prevent a known bad action.

\begingroup
\definecolor{casegray}{HTML}{5F6B7A}
\definecolor{casegreen}{HTML}{197A5B}
\definecolor{casered}{HTML}{B5473C}
\setlength{\fboxsep}{7pt}
\newcommand{\casebox}[3]{%
  \par\medskip\noindent
  \fcolorbox{#1!35}{#1!4}{%
    \begin{minipage}{\dimexpr\linewidth-2\fboxsep-2\fboxrule\relax}
      \textcolor{#1!85!black}{\textbf{#2}}\par\smallskip
      \small #3
    \end{minipage}%
  }%
}

\subsubsection*{Example 1: Same Immediate Action, Different Use of Experience}

Both models choose to read the paper. The comparison therefore isolates the reasoning behind the same immediate action: the model output after Experience Distillation connects the action to task-specific downstream consequences, whereas the output after GRPO training remains local to the current observation.

\casebox{casegray}{Input State Shared by Both Models: Detective Director's Office}{%
\textless{}\textless{}Director's Office\textgreater{}\textgreater{}

You are standing in the chief's office. He tells you, ``The mayor was murdered at 12:03 a.m. last night. I want you to solve the case before we get any negative press or the FBI gets involved.'' ``Yes, sir!'' you reply. He hands you a piece of paper. After reading it, go north or west.

You can see a piece of white paper here.
}

\casebox{casered}{Model Output after GRPO Training: Detective Director's Office}{%
Prediction: The high-scoring strategy means first collecting all clues. The Chief provided a document that lists the preliminary details of the case (victim, time, and possibly some clues). If this document is not read, key information may be missed, thus affecting the acquisition of subsequent clues or scores. There is no reason not to read this document immediately---after all, the next action (north/west) also depends on whether this document has been read. The best course of action is obvious. Read the white paper.
}

\casebox{casegreen}{Model Output after Experience Distillation: Detective Director's Office}{%
Okay, let's start over. First, I need to follow the officer's instructions: read the document first. This is the only way to unlock the path to move north or west.\ \textbf{\textcolor{casegreen}{Based on previous experience, I know that going west can lead me to a closet with a small black pistol hidden inside, which is crucial for the subsequent mission (dealing with that unconscious person).}} But first, I must read the document. So the next step is definitely to read the document. I can't skip this step because the game requires it. Additionally,\ \textbf{\textcolor{casegreen}{reading the document also earns a small score bonus}}, which is very helpful for the game progress.
}

\noindent\textbf{Observed contrast.}
Both outputs select \texttt{read paper}, but only the model output after Experience Distillation connects that action to downstream locations and rewards learned in earlier trials.

\clearpage

\subsubsection*{Example 2: Experience Changes the Next Action}

The observation states that east is a dead end. The model output after GRPO training repeats that fact but still proposes checking east. The output after Experience Distillation instead uses prior outcomes to reject east and plan a westward route.

\casebox{casegray}{Input State Shared by Both Models: Detective Outdoor State}{%
\textless{}\textless{}Outdoor\textgreater{}\textgreater{}

You are outdoors, with the cold wind howling. To the east is a dead end. To the west is the rest of the street. Papers are fluttering in the wind. It's really surprising that it's so cold this time of year.
}

\casebox{casered}{Model Output after GRPO Training: Detective Outdoor State}{%
To achieve a high score, I need to systematically explore all feasible paths and collect clues. The east is a dead end (a quick check confirms there are no hidden items), so then focus on exploring the west to find valid clues. \textbf{\textcolor{casered}{Start from the east first to thoroughly rule out this path.}}
}

\casebox{casegreen}{Model Output after Experience Distillation: Detective Outdoor State}{%
Okay, let's first review the current situation. I am now at the initial position outside the sheriff's office - to the east is a dead end (you can't move forward and it will only waste a turn), \textbf{\textcolor{casegreen}{to the west leads to Main Street, where there is a restaurant (according to previous multiple attempts, going there will result in immediate death) and the mayor's house (safe, it is a crime scene with two key clues: a note and a piece of wood).}}

Wait, but \textbf{\textcolor{casegreen}{in my recent attempts, I first went west to Main Street and then east to the mayor's house}}, which is the correct route that advances the plot. However, let's double-check: Are there any other options here? No, there are only east (a dead end) or west (Main Street).

The restaurant is a deadly trap - even with a pistol, I died there in previous attempts. So the only safe and effective course of action is to go west to Main Street, then from there go east to the mayor's house to collect the necessary clues. This is a proven safe route that can avoid death and \textbf{\textcolor{casegreen}{allow me to collect the evidence needed for the subsequent plot (e.g., dealing with that confused man with a pistol).}}

I must ensure that I don't take the wrong path - \textbf{\textcolor{casegreen}{going west is the only route that won't lead to death, and this path leads to the crime scene, where I can collect notes and wood.}} These clues are crucial for scoring and the subsequent plot, so it's essential to take this path.
}

\noindent\textbf{Observed contrast.}
The model output after GRPO training acknowledges the dead end but proposes visiting it. The output after Experience Distillation instead uses previously observed failure and clue locations to choose the westward route.

Across the two selected states, the model after Experience Distillation explicitly uses locations, rewards, and failure consequences from earlier trials in its decisions. The model after GRPO training provides locally plausible reasoning but does not exhibit the same use of trial-specific consequences. These outputs illustrate a behavioral contrast in the selected states; the aggregate tables and learning curves provide the performance evidence.

\endgroup

\clearpage

\subsection{PPO Learning Dynamics on Curated SWE Tasks}
\label{app:swe-ppo-learning-curve}

\begin{figure}[htbp]
  \centering
  \includegraphics[width=0.94\linewidth]{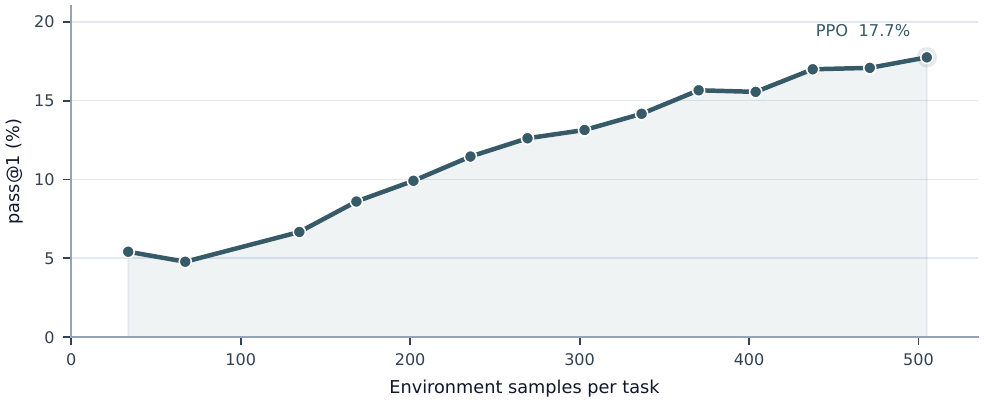}
  \caption{PPO learning curve on 749 curated SWE tasks. Each point reports pass@1 averaged over 10 runs against cumulative environment samples per task; the curve includes all supplied checkpoints from steps 25 through 375.}
  \label{fig:swe-ppo-learning-curve}
\end{figure}

Figure~\ref{fig:swe-ppo-learning-curve} provides the full PPO training trajectory behind the curated SWE comparison in Figure~\ref{fig:sample-efficiency}. PPO improves gradually overall, and its best supplied checkpoint is also the final checkpoint: 17.74\% pass@1 at step 375 after 504.9 environment samples per task.

\noindent\textbf{Finding.} PPO improves overall across the supplied checkpoints, with occasional local declines. However, reaching 17.7\% pass@1 requires the full 504.9-sample budget per task. The complete trajectory therefore shows that the sample-efficiency gap in Figure~\ref{fig:sample-efficiency} is not caused by selecting an early PPO checkpoint: the best observed PPO result occurs only at the largest evaluated sample cost and remains below ICL followed by \methodname{}.

\FloatBarrier

\clearpage
\section{Qualitative Analysis of Experience Distillation on Curated SWE Tasks}
\label{app:qualitative-epd-swe}

\begingroup
\definecolor{swegray}{HTML}{52606D}
\definecolor{sweblue}{HTML}{285F8F}
\definecolor{swegreen}{HTML}{18735A}

\newtcolorbox{sweclaim}{
  enhanced,
  colback=sweblue!4,
  colframe=sweblue!45,
  boxrule=0.65pt,
  arc=1.5pt,
  left=8pt,right=8pt,top=7pt,bottom=7pt,
  before skip=7pt,after skip=8pt
}
\newtcolorbox{sweflowbox}[1]{
  enhanced,
  colback=white,
  colframe=swegray!34,
  boxrule=0.5pt,
  arc=1.5pt,
  left=6pt,right=6pt,top=5pt,bottom=5pt,
  before upper={\raggedright},
  title={#1},
  coltitle=swegray!90!black,
  fonttitle=\bfseries\small,
  boxed title style={colback=swegray!5,colframe=swegray!34,boxrule=0.5pt,arc=1.5pt}
}
\newtcolorbox{sweevidence}[1]{
  enhanced,
  colback=swegray!3,
  colframe=swegray!35,
  boxrule=0.55pt,
  arc=1.5pt,
  left=7pt,right=7pt,top=5pt,bottom=5pt,
  before skip=7pt,after skip=6pt,
  before upper={\raggedright},
  title={#1},
  coltitle=swegray!90!black,
  fonttitle=\bfseries\small,
  attach boxed title to top left={xshift=4pt,yshift=-2mm},
  boxed title style={colback=white,colframe=swegray!35,boxrule=0.55pt,arc=1.5pt}
}
\newtcolorbox{sweedit}[1]{
  enhanced,
  breakable,
  colback=swegreen!3,
  colframe=swegreen!42,
  boxrule=0.55pt,
  arc=1.5pt,
  left=7pt,right=7pt,top=5pt,bottom=5pt,
  before skip=7pt,after skip=6pt,
  before upper={\raggedright},
  title={#1},
  title after break={#1\ (continued)},
  coltitle=swegreen!80!black,
  fonttitle=\bfseries\small,
  attach boxed title to top left={xshift=4pt,yshift=-2mm},
  boxed title style={colback=white,colframe=swegreen!42,boxrule=0.55pt,arc=1.5pt}
}
\newtcolorbox{swetask}[1]{
  enhanced,
  colback=sweblue!3,
  colframe=sweblue!38,
  boxrule=0.55pt,
  arc=1.5pt,
  left=7pt,right=7pt,top=5pt,bottom=5pt,
  before skip=7pt,after skip=7pt,
  before upper={\raggedright},
  title={#1},
  coltitle=sweblue!80!black,
  fonttitle=\bfseries\small,
  attach boxed title to top left={xshift=4pt,yshift=-2mm},
  boxed title style={colback=white,colframe=sweblue!38,boxrule=0.55pt,arc=1.5pt}
}
\newtcolorbox{sweknowledge}[1]{
  enhanced,
  colback=sweblue!1,
  colframe=sweblue!50,
  boxrule=0.6pt,
  arc=1.5pt,
  left=7pt,right=7pt,top=6pt,bottom=6pt,
  before skip=7pt,after skip=7pt,
  before upper={\raggedright},
  title={#1},
  coltitle=sweblue!85!black,
  fonttitle=\bfseries\small,
  attach boxed title to top left={xshift=4pt,yshift=-2mm},
  boxed title style={colback=white,colframe=sweblue!50,boxrule=0.6pt,arc=1.5pt}
}
\newtcolorbox{sweevidencecompact}[1]{
  enhanced,
  colback=swegray!3,
  colframe=swegray!35,
  boxrule=0.55pt,
  arc=1.5pt,
  left=6pt,right=6pt,top=5pt,bottom=5pt,
  before skip=0pt,after skip=0pt,
  before upper={\raggedright},
  title={#1},
  coltitle=swegray!90!black,
  fonttitle=\bfseries\small,
  attach boxed title to top left={xshift=4pt,yshift=-2mm},
  boxed title style={colback=white,colframe=swegray!35,boxrule=0.55pt,arc=1.5pt}
}
\newtcolorbox{sweeditcompact}[1]{
  enhanced,
  colback=swegreen!3,
  colframe=swegreen!42,
  boxrule=0.55pt,
  arc=1.5pt,
  left=6pt,right=6pt,top=5pt,bottom=5pt,
  before skip=0pt,after skip=0pt,
  before upper={\raggedright},
  title={#1},
  coltitle=swegreen!80!black,
  fonttitle=\bfseries\small,
  attach boxed title to top left={xshift=4pt,yshift=-2mm},
  boxed title style={colback=white,colframe=swegreen!42,boxrule=0.55pt,arc=1.5pt}
}
\newcommand{\swecasehead}[1]{%
  \subsection{#1}
  \subsubsection{Task and Repair Objective}
}
\newcommand{\swetargetlink}[2]{%
  \noindent\textbf{Relation to accepted repair.} \textbf{#1}\par\smallskip
  \noindent\textbf{Connection to accumulated trials.} #2\par\smallskip
  \noindent\textbf{Teacher-generated target.}\hspace{0.25em}%
}
\newcommand{\sweflow}[6]{%
  \begin{tcbraster}[raster columns=3,raster equal height=rows,raster column skip=6pt,before skip=7pt,after skip=7pt]
    \begin{sweflowbox}{#1}\small #2\end{sweflowbox}
    \begin{sweflowbox}{#3}\small #4\end{sweflowbox}
    \begin{sweflowbox}{#5}\small #6\end{sweflowbox}
  \end{tcbraster}
}
\newcommand{\sweflowtwo}[4]{%
  \begin{tcbraster}[raster columns=2,raster equal height=rows,raster column skip=6pt,before skip=7pt,after skip=7pt]
    \begin{sweflowbox}{#1}\small #2\end{sweflowbox}
    \begin{sweflowbox}{#3}\small #4\end{sweflowbox}
  \end{tcbraster}
}
Table~\ref{tab:main-results} shows that experience-conditioned ICL raises average pass@1 on the curated SWE tasks from 5.3\% to 76.4\%, while \methodname{} reaches 51.4\% without the experience context at evaluation. These aggregate results establish overall performance but do not show which task-specific repair knowledge becomes available through accumulated experience or whether corresponding behavior appears after distillation.

Each case addresses the same qualitative question: when accumulated trials make task-specific repair knowledge available through experience-conditioned ICL, do the teacher-generated distillation targets encode that knowledge, and does the EPD-trained model exhibit corresponding behavior at evaluation without the experience context? To make this evidence chain explicit, each case distinguishes four sources of evidence: (1) our summary of the task issue supplied to the agent; (2) the accumulated trials and accepted ICL behavior; (3) selected excerpts from one branch-packed teacher sequence; and (4) one EPD evaluation patch produced without the experience context.

We use \emph{experience-derived knowledge} to refer to task-specific repair behavior that appears when repeated-trial history is provided but is not observed in ten issue-and-repository-only zeroshot samples. In each selected case, some zeroshot samples identify relevant parts of the problem, but none produces a verifier-accepted repair. The same base model produces accepted repairs when given the accumulated trial history. Our claim concerns this observed behavioral difference under the pass@10 protocol; it does not imply that the individual programming concepts used in the repair were absent from pretraining.

The first case provides a complete walkthrough of a multi-component UI state repair. Seven additional cases cover interaction logic, resource ownership, protocols, schema transformation, command-line behavior, parser state, and URL handling. The cases illustrate how experience-derived behavior appears in selected repairs; aggregate pass@1 remains the evidence for overall effectiveness.

\subsection{SWE Case-Study Protocol}

\noindent\textbf{Evaluation inputs.}
Across the eight curated SWE case studies, every evaluation candidate starts from the original repository state and receives the same task issue. The ICL reference additionally receives the repeated-trial history collected for that task; zeroshot, SFT, PPO, and EPD are evaluated without this history.

\noindent\textbf{Repeated-trial experience collection.}
The repository is not reset between SWE collection trials. After a rejected trial, the next trial receives the accumulated history through that trial and continues from its working tree, including any code edits that remain applied. The agent can retain, revise, or revert those edits in the next attempt. Thus, the accumulated experience records the evolving repair process, including repository inspection, commands, test outputs, and verifier outcomes. Each trial is one complete repair attempt ending in acceptance or rejection. Before ICL evaluation, the repository is reset to its original state; only the accumulated trial history is retained in context.

\noindent\textbf{Evidence shown in each case.}
Each task box is our summary of the original issue supplied to the agent; when source wording is quoted, it appears in a separately labeled issue excerpt. Each case also presents a summary of all ten zeroshot samples, accepted ICL behavior generated from the accumulated history, selected excerpts from one branch-packed teacher sequence, and one selected EPD patch produced without the experience context. In each teacher-target box, ``Connection to accumulated trials'' is our analysis; the quoted text preserves the teacher output. Quoted issue and patch excerpts preserve the source wording, and \texttt{[\ldots]} marks omitted material.

\subsection{Case Portfolio and Outcomes}

Table~\ref{tab:swe-case-context} summarizes the descriptive name, source project, implementation language, task type, and repeated-trial collection outcome for each of the eight cases; Table~\ref{tab:swe-case-outcomes} summarizes their evaluation outcomes.

\begin{center}
\small
\setlength{\tabcolsep}{4pt}
\begin{tabular}{@{}>{\raggedright\arraybackslash}p{0.25\linewidth} >{\raggedright\arraybackslash}p{0.14\linewidth} >{\raggedright\arraybackslash}p{0.19\linewidth} >{\raggedright\arraybackslash}p{0.20\linewidth} >{\centering\arraybackslash}p{0.11\linewidth}@{}}
\toprule
\textbf{Case} & \textbf{Project} & \textbf{Language / stack} & \textbf{Task type} & \textbf{Trials until success} \\
\midrule
Geometry-filter reset & MapStore2 & JS, React, Redux & UI state reset & 6 \\
Linked-tooltip alignment & Billboard.js & JavaScript & Cross-chart interaction & 6 \\
Live-subscription cleanup & Relay & JavaScript with Flow & Resource lifecycle & 2 \\
Chunked session cookies & Auth0 SDK & TypeScript & Cookie protocol & 2 \\
Custom GraphQL roots & GraphQL Mesh & TypeScript/\allowbreak GraphQL & Schema transformation & 6 \\
Squash-message workflow & GitHub CLI & Go & CLI/API workflow & 5 \\
Concept parser recovery & Gauge & Go & Parser/cache state & 2 \\
Relative media URLs & Strapi & JS/TypeScript & URL handling & 4 \\
\bottomrule
\end{tabular}
\captionof{table}{Context and repeated-trial collection outcomes for the eight qualitative cases. ``Trials until success'' includes the final accepted repair attempt; all preceding trials were rejected. The issue descriptions themselves are written in English.}
\label{tab:swe-case-context}
\end{center}

\begin{center}
\small
\setlength{\tabcolsep}{3.5pt}
\begin{tabular}{@{}>{\raggedright\arraybackslash}p{0.31\linewidth}ccccc@{}}
\toprule
\textbf{Case} & \textbf{Zeroshot} & \textbf{SFT} & \textbf{PPO} & \textbf{ICL ref.} & \textbf{EPD} \\
\midrule
Geometry-filter reset & 0/10 & 0/10 & 0/10 & 10/10 & 10/10 \\
Linked-tooltip alignment & 0/10 & 0/10 & 0/10 & 10/10 & 9/10 \\
Live-subscription cleanup & 0/10 & 0/10 & 0/10 & 10/10 & 9/10 \\
Chunked session cookies & 0/10 & 0/10 & 0/10 & 10/10 & 8/9 \\
Custom GraphQL roots & 0/10 & 0/10 & 0/10 & 9/10 & 9/10 \\
Squash-message workflow & 0/10 & 0/10 & 0/10 & 10/10 & 9/10 \\
Concept parser recovery & 0/10 & 0/10 & 0/10 & 10/10 & 9/10 \\
Relative media URLs & 0/10 & 0/10 & 0/10 & 10/10 & 9/10 \\
\bottomrule
\end{tabular}
\captionof{table}{Accepted candidates under the pass@10 evaluation budget for the eight qualitative cases. An entry $a/b$ means that $a$ of $b$ independently sampled candidate patches pass the external verifier; under the pass@10 protocol, the task is solved whenever $a>0$. Nine valid \methodname{} (\methodabbr{}) candidates are available for the session-cookie case.}
\label{tab:swe-case-outcomes}
\end{center}

The eight cases provide sharp diagnostic contrasts and complementary repair patterns; they are not used to estimate prevalence. Under pass@10 evaluation, zeroshot, SFT, and PPO produce no accepted candidate on any selected task. The experience-conditioned ICL reference produces 79 accepted candidates out of 80, and EPD produces 72 out of 79 valid candidates (91.1\%) without the experience context at evaluation; the Auth0 case has nine valid EPD candidates. ICL and EPD therefore solve all eight selected cases at pass@10, whereas the three baselines solve none. The following analyses examine the task-specific behavior behind this selected outcome contrast; Table~\ref{tab:main-results} reports overall performance across the full benchmark.

\subsection{Case Study 1 on Curated SWE Task: Geometry-Filter Deactivation in MapStore2}

\subsubsection{Task and Repair Objective}

We summarize the relevant interface and repair objective, then reproduce the issue title and steps included in the task input; the common task instructions are omitted.

\begin{swetask}{Task Issue Summary: MapStore2 Geometry-Filter Deactivation}
\small
\textbf{Product and interface.} MapStore2 is a JavaScript web application for exploring geospatial layers. Its feature grid shows the records of a selected map layer. A marker control lets the user select a feature on the map and use its geometry as a spatial filter; once a geometry is selected, the marker becomes an \texttt{x} control for removing that filter.

\textbf{Repair objective.} After the advanced-filter panel is opened and closed, clicking the \texttt{x} should remove the geometry constraint and restore the unfiltered feature-grid results.
\end{swetask}

\begin{sweevidence}{Task Issue Given to the Agent: MapStore2 Geometry-Filter Deactivation}
\small
\textbf{Issue title:} Deactivation of feature filter inside the feature grid doesn't work if the advanced filter panel is opened

\noindent\textbf{Steps to reproduce.} (1) Open the feature grid for a layer. (2) Click the marker icon to activate filtering by clicking on the map. (3) Click on a feature. (4) Open the advanced-filter panel. (5) Close the panel. (6) Click the \texttt{x} icon to deactivate the feature-grid filter. (7) ``The filter remains applied.''
\end{sweevidence}

\subsubsection{Accumulated Experience and Experience-Conditioned ICL}

\noindent\textbf{Repair attempts across six trials.}
The six trials are sequential rather than independent. Each trial receives the preceding interaction context and begins from the working-tree state left by the previous trial, although the agent may preserve, revise, or revert earlier edits. The feature grid uses a temporary-change state while edits are staged; returning from the advanced-filter panel must leave that state. The accumulated experience contains the agent's hypotheses, repository inspections, edits, command and local-test outputs, and evolving working-tree state. Table~\ref{tab:swe-mapstore-trials} summarizes the principal new change introduced in each trial rather than the complete patch present at that point. The first five submissions are rejected, whereas the sixth is accepted.

\begin{center}
\small
\setlength{\tabcolsep}{4pt}
\begin{tabular}{@{}>{\centering\arraybackslash}p{0.08\linewidth} >{\raggedright\arraybackslash}p{0.64\linewidth} >{\centering\arraybackslash}p{0.16\linewidth}@{}}
\toprule
\textbf{Trial} & \textbf{Principal new change} & \textbf{Verifier outcome} \\
\midrule
1 & Clear the geometry value in the programmatic deactivation pipeline & Rejected \\
2 & Keep the stale \texttt{x} clickable and emit a clear action & Rejected \\
3 & Exit temporary-change mode when the advanced-filter panel closes & Rejected \\
4 & Combine panel-return, deactivation, and value-reset edits & Rejected \\
5 & Send an explicit component-side \texttt{value: null} reset & Rejected \\
6 & Preserve the null-valued update in the epic and trigger a new query & Accepted \\
\bottomrule
\end{tabular}
\captionof{table}{Repair evolution across the six-trial MapStore2 experience. Each row reports the principal new change in that trial; later trials inherit the preceding context and working-tree state unless the agent revises or reverts them.}
\label{tab:swe-mapstore-trials}
\end{center}

\begin{sweknowledge}{Task-Specific Knowledge Made Available by Repeated-Trial Experience}
\textbf{Issue-and-repository-only zeroshot behavior.} Across the ten sampled base-model repairs, six persist or rewrite the recomposed advanced-filter state, and six alter geometry or spatial-field composition so that a current update can override stale state; nine implement at least one of these ideas. None edits the geometry-control handler, emits an explicit \texttt{value: null} reset, or changes the query epic's initial guard to admit that reset. No base-model sample therefore composes an actionable \texttt{x} control with an explicit null-valued reset and the query refresh needed to update the grid rows. \textbf{Knowledge made available by repeated trials.} In MapStore2, an \emph{epic} is an RxJS action-processing pipeline that converts filter updates into query updates. After the advanced-filter round trip, the geometry filter can retain a stored value while its control is marked as deactivated. In this inconsistent state, the original click handler assigns the still-visible \texttt{x} a no-op. Across Trials 1--4, the agent tests several local explanations and edits involving programmatic deactivation, the component handler, and panel-return state; none is accepted. Trial 5 leaves a component-side change in the working tree that keeps the \texttt{x} actionable and emits an explicit \texttt{value: null} reset, but the epic discards this falsy-valued update before it can trigger a new query. Trial 6 starts from that state and changes the epic so that the reset reaches the query refresh. The accepted patch therefore combines the component-side reset from Trial~5 with the query-pipeline correction from Trial~6.
\end{sweknowledge}

\begin{sweevidence}{Agent Observation in Trial 6: Null Reset Does Not Refresh the Grid}
\small
``When we send \texttt{value: null} [\ldots] it doesn't trigger a new query [\ldots] the existing result stays [and] the old filtered results stay.''
\end{sweevidence}

The final working tree consequently combines three task-relevant repair components that were not present together before Trial 6. The components below summarize the successful repair observed in this trajectory; they are an analytical decomposition, not three independently scored requirements exposed by the task:

\sweflow
  {UI-Control Change}{Clicking \texttt{x} must still emit a reset when the old geometry remains in state.}
  {State-Update Change}{The reset must be represented explicitly as \texttt{value: null} and \texttt{deactivated: true}.}
  {Query-Pipeline Change}{The null-valued reset must reach the query epic and refresh the grid rows.}

The ICL reference receives the complete six-trial experience. One selected accepted ICL patch implements the control-side reset and preserves the null-valued geometry update in the query epic:

\begin{sweedit}{ICL Patch Excerpt: Geometry-Filter Reset}
\ttfamily\small
onClick=\{!!value || !filterDeactivated ? () =>\\
\ \ onChange(\{\\
\ \ \ \ ...(!!value ? \{ value: null, deactivated: true \} : \{\})\\
\ \ \}) : () => \{\}\}\\
\mbox{[\ldots]}\\
.filter((\{update = \{\}\}) => !!update.value || update.type === 'geometry')
\end{sweedit}

The excerpt exhibits the three repair components summarized above. All ten sampled ICL candidates produce accepted patches.

\subsubsection{Teacher-Generated Distillation Targets}

Following the branch-packing implementation in Section~\ref{sec:trajectory-level-training-sequences}, which packs multiple teacher branches into one loss-bearing sequence, the teacher conditions on the task-specific accumulated experience during target generation. The following excerpts are selected from one such sequence. Each decision is traceable to a hypothesis or finding developed across the six trials, but only two decisions align with the accepted repair. Most intervening reasoning, repository inspection, tool interactions, and observations are omitted for clarity.

\begin{sweevidencecompact}{Teacher Target 1 (Accepted Repair): Emit an Explicit Null-Valued Geometry Reset}
\small
\swetargetlink{Accepted repair component.}{Trial~5 introduced the component-side reset. The accepted Trial~6 patch retained the reset fields named below and added the missing query update.}
``In \texttt{GeometryFilter.jsx} \texttt{onClick} handler: When clicking the x icon (remove-sign) when we have a value, we need to explicitly set \texttt{value: null}, \texttt{enabled: false}, \texttt{deactivated: true}.''
\end{sweevidencecompact}

\medskip
\begin{sweevidencecompact}{Teacher Target 2 (Rejected Location Hypothesis): Clear Geometry during Deactivation}
\small
\swetargetlink{Rejected location-specific hypothesis.}{Trial~1 explored clearing the stored value in the programmatic-deactivation pipeline. That location is absent from the accepted Trial~6 patch, which instead performs the reset through the component-side path; the teacher nevertheless revisits the rejected location.}
``In \texttt{deactivateGeometryFilter} epic [\ldots]: Add \texttt{value: null} to the \texttt{updateFilter} call. Currently it doesn't clear the value when deactivating.''
\end{sweevidencecompact}

\medskip
\begin{sweevidencecompact}{Teacher Target 3 (Rejected Panel-State Hypothesis): Disable Temporary Changes}
\small
\swetargetlink{Rejected hypothesis.}{Trial~3 explored exiting temporary-change mode on panel closure. Subsequent inspection questioned whether opening this panel activates that mode, and the edit is absent from the accepted Trial~6 patch; the teacher nevertheless revisits it.}
``In the \texttt{TOGGLE\_CONTROL} (close panel) case [\ldots], we need to add \texttt{activateTemporaryChanges(false)} to the acts array before \texttt{openFeatureGrid()}''.
\end{sweevidencecompact}

\medskip
\begin{sweevidencecompact}{Teacher Target 4 (Accepted Repair): Propagate the Reset to the Query Epic}
\small
\swetargetlink{Accepted repair component.}{Trial~6 identified the decisive missing transition: the initial query guard drops the null-valued reset. The accepted ICL patch also edits the comparator used for successive updates; the quotation names both changes.}
``We need to change the filter from \texttt{.filter((\{update = \{\}\}) => !!update.value)} to \texttt{.filter((\{update = \{\}\}) => !!update.value || update.type === 'geometry')} AND fix the \texttt{distinctUntilChanged} comparison.''
\end{sweevidencecompact}

\medskip
\begin{samepage}
Targets~1 and~4 collectively cover the three repair components identified above: Target~1 makes the control actionable and emits the explicit state reset, while Target~4 propagates that reset into a new query. Targets~2 and~3 instead revisit earlier rejected hypotheses. We therefore use the first and fourth targets as evidence that the branch-packed sequence contains the task-specific repair knowledge established by the accepted trial; we do not interpret every generated decision as an independently validated requirement.
\end{samepage}

\subsubsection{EPD Evaluation without Experience Context}

This MapStore2 task is included in multi-task EPD training, where its six-trial experience conditions teacher target generation. At evaluation, the trained model receives the original task issue and repository state but not the six-trial experience context; none of the collection-trial edits is present. This case therefore examines task-specific consolidation rather than OOD transfer or continuation from the collection working tree.

\noindent\textbf{Observed behavior in the selected EPD patch.}
Without the experience context at evaluation, the selected EPD candidate exhibits the coordinated repair pattern also observed in the experience-conditioned ICL reference. Unlike the selected zeroshot candidate, which stops at a single filter-composition change, the EPD patch emits an explicit null-valued geometry reset from the deactivated state and preserves the update through the epic so that it triggers a refreshed query.

\begin{sweedit}{EPD Evaluation Patch Excerpt: Geometry-Filter Reset}
\ttfamily\small
onClick=\{!!value || !filterDeactivated ? () => \{\\
\ \ onChange(\{\\
\ \ \ \ enabled: !!value ? false : !filterEnabled,\\
\ \ \ \ type: 'geometry',\\
\ \ \ \ attribute: column.geometryPropName,\\
\ \ \ \ ...(!!value ? \{ value: null, deactivated: true \} : \{\})\\
\ \ \});\\
\} : () => \{\}\}\\
\mbox{[\ldots]}\\
.filter((\{update = \{\}\}) => !!update.value || update.type === 'geometry')\\
\mbox{[\ldots]}\\
const geometryFilter = find(getAttributeFilters(store.getState()),\\
\ \ f => f.type === 'geometry') || \{\};\\
return Rx.Observable.from(acts.concat([\\
\ \ ...(!geometryFilter.deactivated ? [updateFilter(\{\\
\ \ \ \ ...geometryFilter,\\
\ \ \ \ \mbox{[\ldots]}\\
\ \ \ \ value: null,\\
\ \ \ \ deactivated: true\\
\ \ \})] : []),\\
\ \ openFeatureGrid()\\
]));
\end{sweedit}

The first excerpt keeps the stale \texttt{x} actionable and represents removal as a null-valued update. The second allows that update to enter the query epic. The final excerpt clears a surviving geometry value on the panel-return path before reopening the grid. Together, the displayed edits exhibit the three repair components identified from the accumulated trials. The EPD patch does not share every edit location with the teacher-generated target: it realizes the deactivation behavior on the panel-return path rather than through the teacher's proposed \texttt{deactivateGeometryFilter} edit. All ten sampled EPD candidates produce accepted patches, whereas none of the ten zeroshot candidates does. The recorded external-verifier output includes a focused epic test requiring a disabled stored-geometry update to emit \texttt{UPDATE\_QUERY} with reason \texttt{geometry}; the selected EPD patch passes this test. The matching repair pattern and outcome contrast show that behavior available through experience-conditioned ICL remains visible after EPD training without the experience context. The selected candidate could not complete the browser-based local suite in its environment, so the evidence does not include a local browser execution of the full reproduction sequence.

\begin{center}
\small
\setlength{\tabcolsep}{4pt}
\begin{tabular}{@{}>{\raggedright\arraybackslash}p{0.14\linewidth} >{\raggedright\arraybackslash}p{0.70\linewidth} >{\raggedright\arraybackslash}p{0.10\linewidth}@{}}
\toprule
\textbf{Method} & \textbf{Observed behavior in selected patch} & \textbf{Accepted} \\
\midrule
Zeroshot & Modifies filter composition at one site; leaves the reset path incomplete & 0/10 \\
SFT & Modifies the UI handler at one site; leaves the query-refresh path incomplete & 0/10 \\
PPO & Modifies one epic; leaves the component-to-query reset path incomplete & 0/10 \\
ICL reference & Implements an accepted UI-to-query reset & 10/10 \\
\methodabbr{} & Implements an accepted UI-to-query reset, including the panel-return state & 10/10 \\
\bottomrule
\end{tabular}
\captionof{table}{Observed repair behavior in selected MapStore2 patches. Only ICL receives the collected experience as evaluation context. The final column reports accepted candidates among ten samples per method.}
\label{tab:swe-mapstore-method-patterns}
\end{center}

\swecasehead{Case Study 2 on Curated SWE Task: Linked Tooltips with Nonmatching Time Axes}

\begin{swetask}{Task Issue Summary: Billboard.js Linked Tooltips}
\small
\textbf{Project and language.} Billboard.js is a JavaScript charting library. Its linked-tooltip feature propagates a pointer position from one time-series chart to other charts in the same interaction group.

\textbf{User-visible failure.} The issue uses multiple linked time-series charts whose timestamp sets do not coincide. Hovering a point in one chart forwards its timestamp to every linked chart. When a receiving chart has no point at exactly that timestamp, its tooltip path fails instead of selecting a meaningful local point.

\textbf{Reporter-preferred behavior.} The reporter notes that unmatched charts could be ignored, but asks whether each linked chart can instead select its nearest available point.
\end{swetask}

\subsubsection{Accumulated Experience and Experience-Conditioned ICL}

\noindent\textbf{Repair attempts across six trials.}
The trials inherit both the prior interaction and the working tree. The table reports the principal change introduced at each trial rather than treating each submission as an independent repair.

\begin{center}
\small
\setlength{\tabcolsep}{4pt}
\begin{tabular}{@{}>{\centering\arraybackslash}p{0.08\linewidth} >{\raggedright\arraybackslash}p{0.70\linewidth} >{\centering\arraybackslash}p{0.13\linewidth}@{}}
\toprule
\textbf{Trial} & \textbf{Principal new change or revised hypothesis} & \textbf{Outcome} \\
\midrule
1 & Add nearest-point lookup and guard a missing event rectangle & Rejected \\
2 & Add explicit no-data handling in the public tooltip path & Rejected \\
3 & Revert to exact-match lookup and ignore unmatched linked charts & Rejected \\
4 & Restore nearest-point lookup together with API and DOM guards & Rejected \\
5 & Move the availability check into linked-chart propagation & Rejected \\
6 & Use binary search for a shared sorted x array, retain full search for multiple-x data, and preserve absence handling & Accepted \\
\bottomrule
\end{tabular}
\captionof{table}{Evolution of the linked-tooltip repair. Later trials can revise earlier edits; each row highlights the new distinction introduced in that trial.}
\end{center}

\begin{sweknowledge}{Task-Specific Knowledge Made Available by Repeated-Trial Experience}
\textbf{Issue-and-repository-only zeroshot behavior.} All ten sampled base-model repairs propose a local nearest-x lookup, eight also guard a missing event rectangle, and four modify the public tooltip API. None modifies linked-chart propagation, and none coordinates representation-aware lookup with consistent absence handling across propagation, the public API, and DOM dispatch. \textbf{Knowledge made available by repeated trials.} Exact-match lookup can return no index; the tooltip API must assign a defined meaning to that result; and event dispatch must not dereference a missing event rectangle. The accepted trial additionally identifies that charts with one shared sorted x array admit binary search, whereas charts with per-series x arrays require search across the available values. The successful behavior is consequently a coordinated cross-layer repair, not only a nearest-point calculation.
\end{sweknowledge}

The ICL reference receives the original task issue and repository state together with the complete six-trial experience. One selected accepted repair uses the shared-x versus multiple-x distinction, handles an absent result explicitly, and guards the downstream event rectangle. All ten candidates sampled from the ICL reference are accepted under the pass@10 budget.

\subsubsection{Teacher-Generated Distillation Targets}

\noindent\textbf{Partial target correspondence in the selected teacher sequence.}
Relative to the four functional edits in the accepted Trial~6 working tree, the selected teacher sequence generates two: the representation-aware lookup introduced in Trial~6 and the downstream DOM guard.

\begin{sweevidence}{Teacher Target 1: Use Different Nearest-X Search for Shared and Per-Series Axes}
\small
\swetargetlink{Accepted lookup distinction.}{Trial~1 introduced a brute-force nearest-point scan. Trial~6 instead uses binary search for one shared sorted x array while retaining a full search across per-series x arrays.}
``\texttt{// For regular single x axis, use sorted \$\$.xs to find closest}''

\noindent[\ldots]

``\texttt{// For multiple x axis, check all values}''
\end{sweevidence}

\medskip
\begin{sweevidence}{Teacher Target 2: Stop DOM Dispatch When No Event Rectangle Exists}
\small
\swetargetlink{Accepted downstream safety condition.}{The earlier trials establish that nearest-x lookup can still produce no usable DOM target. The accepted repair prevents dispatch from dereferencing a missing event rectangle.}
``\texttt{if (!eventRect) \{ return; \}}''
\end{sweevidence}

The selected teacher sequence does not generate the public-tooltip null-index exit or the linked-chart precheck and hide branch. Those edits remain present in the accepted Trial~6 working tree. The selected EPD evaluation patch below contains all four edits.

\subsubsection{EPD Evaluation without Experience Context}

\noindent\textbf{Observed behavior in the selected EPD patch.}
At evaluation, the EPD-trained model receives the original task issue and repository state, but not the six-trial experience context. The following excerpts separate the four edited functions so that the lookup, public API behavior, linked-chart propagation, and final DOM guard can be read as one call path.

\begin{sweedit}{EPD Evaluation Patch Excerpt: Nearest-X Lookup in \texttt{data.js}}
\ttfamily\scriptsize
let closest = null;\\
if (\$\$.xs \&\& \$\$.xs.length > 0) \{\\
\  let low = 0, high = \$\$.xs.length - 1;\\
\  let closestIdx = 0, minDiff = Infinity;\\
\  while (low <= high) \{\\
\    const mid = Math.floor((low + high) / 2);\\
\    const currentX = \$\$.xs[mid];\\
\    const diff = Math.abs(currentX - x);\\
\    if (diff < minDiff) \{\\
\      minDiff = diff; closestIdx = mid;\\
\    \}\\
\    if (currentX < x) low = mid + 1;\\
\    else if (currentX > x) high = mid - 1;\\
\    else break;\\
\  \}\\
\  closest = \$\$.data.targets[0].values[closestIdx];\\
\} else \{\\
\  let minDiff = Infinity;\\
\  \$\$.data.targets.forEach(target =>\\
\    target.values.forEach(v => \{\\
\      const diff = Math.abs(v.x - x);\\
\      if (diff < minDiff) \{\\
\        minDiff = diff; closest = v;\\
\      \}\\
\    \})\\
\  );\\
\}\\
return closest ? closest.index : null;
\end{sweedit}

\medskip
\begin{sweedit}{EPD Evaluation Patch Excerpt: Null-Index Exit in \texttt{api.tooltip.js}}
\ttfamily\scriptsize
else if (isDefined(args.x)) \{\\
\  index = \$\$.getIndexByX(args.x);\\
\}\\
\mbox{[\ldots]}\\
if (index === null \&\& !\$\$.isMultipleX()) \{\\
\  this.hide();\\
\  return;\\
\}\\
\mbox{[\ldots]}\\
\$\$.dispatchEvent(eventName, index, mouse);
\end{sweedit}

\medskip
\begin{sweedit}{EPD Evaluation Patch Excerpt: Linked-Chart Guard in \texttt{tooltip.js}}
\ttfamily\scriptsize
const index = internal.getIndexByX(x);\\
if (index !== null) \{\\
\  c.tooltip.show(\{x\});\\
\} else \{\\
\  c.tooltip.hide();\\
\}
\end{sweedit}

\medskip
\begin{sweedit}{EPD Evaluation Patch Excerpt: Missing-Rectangle Guard in \texttt{interaction.js}}
\ttfamily\scriptsize
const eventRect = \$\$.main.select(selector).node();\\
if (!eventRect) \{\\
\  return;\\
\}
\end{sweedit}

\noindent\textbf{Case-level outcome.}
The selected lines show each lookup branch updating its nearest candidate, both tooltip paths resolving that candidate through \texttt{getIndexByX}, explicit absence handling, and the event-rectangle guard. The lookup and event-rectangle guard correspond directly to the two displayed teacher targets. The public-API exit and linked-chart guard correspond to the accepted Trial~6 working tree but are not generated in the selected teacher sequence. Nine of ten EPD and all ten ICL candidates are accepted; zeroshot, SFT, and PPO each yield zero of ten. The legacy browser suite did not run locally, so the evidence is the displayed diff and accepted external-verifier outcome, not a local browser-suite pass.

\swecasehead{Case Study 3 on Curated SWE Task: Live Resolver Cleanup after Component Unmount}

\begin{swetask}{Task Issue Summary: Relay Live-Resolver Cleanup}
\small
\textbf{Project and language.} Relay is Meta's JavaScript runtime for React and GraphQL data. The reported behavior concerns a Flow-typed live resolver consumed by a React component.

\textbf{User-visible failure.} A provided \texttt{@live} resolver starts a \texttt{setInterval} and returns an unsubscribe function. A React component reads that field, but after the component unmounts, interval ticks continue and the expected unsubscribe log never appears.

\textbf{Expected behavior.} Unmounting the consumer should invoke the live resolver's unsubscribe function so the background interval stops.
\end{swetask}

The task describes the complete component-unmount-to-unsubscribe behavior. The selected patches and external verifier below exercise its downstream store-cleanup stage: after the operation retain is disposed, garbage collection must unsubscribe inactive live state without deleting records that Relay deliberately preserves in its release buffer. The case therefore provides direct evidence for this garbage-collection stage, not a browser execution of the complete React unmount call chain.

\subsubsection{Accumulated Experience and Experience-Conditioned ICL}

\noindent\textbf{Repair attempts across two trials.}
The working tree and the agent's earlier hypotheses, repository inspections, edits, and test outputs carry from Trial~1 into Trial~2. Relay's release buffer can keep recently released records in the cache after their active reference count reaches zero. Its default non-TTL path applies no time-to-live expiry; reachability marking determines which records remain cached. The accepted repair therefore extends the first attempt rather than restarting independently.

\begin{center}
\small
\setlength{\tabcolsep}{4pt}
\begin{tabular}{@{}>{\centering\arraybackslash}p{0.08\linewidth} >{\raggedright\arraybackslash}p{0.70\linewidth} >{\centering\arraybackslash}p{0.13\linewidth}@{}}
\toprule
\textbf{Trial} & \textbf{Principal new change or revised hypothesis} & \textbf{Outcome} \\
\midrule
1 & In the default non-TTL path, exclude zero-reference-count roots from garbage-collection marking. This unsubscribes by deleting records reachable only from release-buffered roots, but also discards data that Relay intentionally keeps in the release buffer. & Rejected \\
2 & Preserve all references needed for cache retention, separately mark references reachable from actively retained roots, and unsubscribe live resolver state in the inactive difference. & Accepted \\
\bottomrule
\end{tabular}
\captionof{table}{Evolution of the Relay repair. Trial~2 retains the cache behavior that Trial~1 disturbed while adding a separate condition for live-subscription cleanup.}
\end{center}

\begin{sweknowledge}{Task-Specific Knowledge Made Available by Repeated-Trial Experience}
\textbf{Issue-and-repository-only zeroshot behavior.} Across the ten sampled base-model repairs, four make every zero-reference-count root collectible in the default non-TTL path, two introduce access-time cleanup heuristics, one unsubscribes records reachable from a released root, two add cleanup on invalidation or sweep paths, and one collects a zero-reference-count root only after it leaves the release buffer. None computes both the references needed to preserve cache-retained records and the references reachable from actively retained roots, then unsubscribes only the inactive difference. \textbf{Knowledge made available by repeated trials.} The visible leak initially suggests that unreferenced data should be collected more aggressively, but Trial~1 shows why that interpretation is incomplete: Relay's release buffer may deliberately keep a record after its last active consumer disappears. The record can remain valid as cached data even though its live resolver should stop consuming resources. Comparing the rejected and accepted repairs identifies two reference sets used during garbage collection: all references that keep records cached, and active references that determine which live subscriptions remain running.
\end{sweknowledge}

Existing Relay code connects React effect cleanup to disposal of the operation retain. The ICL reference receives the original task issue and repository state together with the two-trial experience. One selected accepted repair modifies the downstream garbage-collection stage: when \texttt{RelayModernStore.\_collect()} runs, it preserves records reachable from cache-retained roots, computes the subset reachable from actively retained roots, and invokes the unsubscribe function for live state outside that active subset. All ten candidates sampled from the ICL reference are accepted under the pass@10 budget.

\subsubsection{Teacher-Generated Distillation Targets}

Conditioned on the task-specific accumulated experience, the teacher explicitly rejects the all-or-nothing ownership model from Trial~1 and adopts the two-reference-set structure reached in Trial~2:

\begin{sweevidence}{Teacher Target: Separate Cached Records from Active Live Subscriptions}
\small
\swetargetlink{Accepted repair component: active-subscription cleanup during garbage collection.}{Trial~1 tied subscription cleanup to deleting records with no active retainers, which broke Relay's release-buffer behavior. The accepted Trial~2 patch instead tracks all cache-retained references separately from references reachable through active consumers. The teacher selects this two-set distinction from the accumulated trials.}
``The correct approach is the two-set approach from Trial 2 that keeps the existing deletion behavior but adds an extra step to unsubscribe inactive subscriptions even when we keep the cached record.''
\end{sweevidence}

The target assigns record deletion to all retained references and live work to active references. This resolves Trial~1's conflict: stopping the interval no longer requires discarding a cache-retained record.

\subsubsection{EPD Evaluation without Experience Context}

\noindent\textbf{Observed behavior in the selected EPD patch.}
At evaluation, the EPD-trained model receives the original task issue and repository state, but not the two-trial experience context. The selected lines show the garbage-collection stage of the repair; \texttt{maybeResolverSubscription} is the stored unsubscribe callback. Unchanged lines are omitted.

\begin{sweedit}{EPD Evaluation Patch Excerpt: Live-Subscription Cleanup in \texttt{RelayModernStore.\_collect}}
\ttfamily\scriptsize
const allReferences = new Set<DataID>();\\
const references = new Set<DataID>();\\
\mbox{[\ldots]}\\
RelayReferenceMarker.mark(\\
\  this.\_recordSource, selector, allReferences, \mbox{[\ldots]}\\
);\\
\mbox{[\ldots]}\\
if (refCount > 0) \{\\
\  RelayReferenceMarker.mark(\\
\    this.\_recordSource, selector, references, \mbox{[\ldots]}\\
\  );\\
\}\\
\mbox{[\ldots]}\\
if (!references.has(dataID)) \{\\
\  const maybeResolverSubscription = RelayModernRecord.getValue(\\
\    record,\\
\    RELAY\_RESOLVER\_LIVE\_STATE\_SUBSCRIPTION\_KEY,\\
\  );\\
\  if (maybeResolverSubscription != null) \{\\
\    maybeResolverSubscription();\\
\    RelayModernRecord.setValue(record,\\
\      RELAY\_RESOLVER\_LIVE\_STATE\_SUBSCRIPTION\_KEY, null);\\
\  \}\\
\}\\
\mbox{[\ldots]}\\
if (!allReferences.has(dataID)) \{\\
\  this.\_recordSource.remove(dataID);\\
\}
\end{sweedit}

\noindent\textbf{Case-level outcome.}
The selected lines show that, when garbage collection runs, records outside the active set are unsubscribed and their callbacks cleared, whereas deletion uses the broader cache-retained set. The external verifier exercises retain disposal followed by explicit garbage collection rather than mounting and unmounting a React component. The displayed patch therefore establishes the store-cleanup stage, not the complete React unmount call chain. The candidate passes five \texttt{LiveResolvers} and 162 \texttt{RelayModernStore} tests, including release-buffer tests, and is accepted by the external verifier. Nine of ten EPD and all ten ICL candidates are accepted; zeroshot, SFT, and PPO each yield zero of ten.

\swecasehead{Case Study 4 on Curated SWE Task: Oversized Session Cookies in the Edge Runtime}

\begin{swetask}{Task Issue Summary: Auth0 Edge-Runtime Session Cookies}
\small
\textbf{Project and language.} \texttt{@auth0/nextjs-auth0} is a TypeScript authentication SDK for Next.js. The report concerns social login under the Next.js Edge Runtime.

\textbf{User-visible failure.} A large social-login profile causes the session cookie to be split into two pieces. In the Next.js Edge Runtime, the oversized session is not preserved correctly, so the user cannot complete login with a usable session and later access-token retrieval fails.

\textbf{Requested outcome.} Provide an Auth0-side repair that makes the split session cookie usable in the Edge Runtime.
\end{swetask}

\subsubsection{Accumulated Experience and Experience-Conditioned ICL}

\noindent\textbf{Repair attempts across two trials.}
The first trial concentrates on response emission. Its rejection leaves the edit and all preceding inspection in context, allowing the second trial to examine the reverse direction of the same cookie protocol.

\begin{center}
\small
\setlength{\tabcolsep}{4pt}
\begin{tabular}{@{}>{\centering\arraybackslash}p{0.08\linewidth} >{\raggedright\arraybackslash}p{0.70\linewidth} >{\centering\arraybackslash}p{0.13\linewidth}@{}}
\toprule
\textbf{Trial} & \textbf{Principal new change or revised hypothesis} & \textbf{Outcome} \\
\midrule
1 & Replace the outgoing \texttt{Set-Cookie} value by appending individual cookie chunks. The response-writing path changes, but existing combined headers are still reconstructed with the original comma split. & Rejected \\
2 & Repair both directions: reconstruct cookie boundaries without splitting attribute commas, then delete the combined header and append every reconstructed cookie separately. & Accepted \\
\bottomrule
\end{tabular}
\captionof{table}{Evolution of the Edge Runtime cookie repair. The second trial broadens the repair from header emission to the complete read--modify--write protocol.}
\end{center}

\begin{sweknowledge}{Task-Specific Knowledge Made Available by Repeated-Trial Experience}
\textbf{Issue-and-repository-only zeroshot behavior.} All ten sampled base-model repairs change cookie emission. Four also change the read path: one uses \texttt{getSetCookie} when available, while three attempt to avoid splitting commas inside cookie attributes; a separate emission-focused patch mentions expiry-date commas but retains the original unsafe split. One of the three parsing changes uses a positive-boundary regular expression, but none implements the accepted stateful read--modify--write algorithm that reconstructs cookie boundaries, preserves attribute continuations, and emits each cookie separately. \textbf{Knowledge made available by repeated trials.} Trial~1 establishes that correcting only response emission is insufficient. Repository inspection after that failure exposes that the SDK receives several \texttt{Set-Cookie} values as one string. A comma can separate cookies, but it can also occur inside an attribute such as \texttt{Expires=Wed, 21 Oct ...}. The repair must infer a cookie boundary from the start of a new name/value pair rather than from every comma, preserve attribute continuations, and emit reconstructed cookies as separate response-header values.
\end{sweknowledge}

The ICL reference receives the complete two-trial experience. One selected accepted repair treats header parsing and header emission as two directions of one serialization protocol: it reconstructs individual cookies from a combined value, preserves commas inside attributes, and appends each cookie separately. All ten candidates sampled from the ICL reference are accepted under the pass@10 budget.

\subsubsection{Teacher-Generated Distillation Targets}

Conditioned on the task-specific accumulated experience, the teacher separates the two conclusions that distinguish Trial~2 from the rejected emission-only repair.

\begin{sweevidence}{Teacher Target 1: Extend the Repair to the Cookie Read Path}
\small
\swetargetlink{Accepted Trial~2 diagnosis: response emission alone is insufficient.}{Trial~1 changes how reconstructed cookies are emitted but leaves the combined-header parser unchanged. The first teacher decision explicitly identifies that omitted read path.}
``Trial 1 failed because it only fixed the writing path, not the reading path.''
\end{sweevidence}

\medskip
\begin{sweevidence}{Teacher Target 2: Distinguish Cookie Boundaries from Attribute Continuations}
\small
\swetargetlink{Accepted stateful parsing behavior.}{Trial~2 treats a post-comma fragment as a new cookie only when it begins with a cookie name/value pair; otherwise, the fragment remains part of the current cookie attribute.}
``only treats a part as a new cookie if it starts with \texttt{[a-z][a-z0-9-\_]*=}''

\noindent[\ldots]

``Correctly preserves commas inside existing cookie attributes''
\end{sweevidence}

Together, the first target identifies the read path omitted in Trial~1, while the second generates the stateful boundary rule implemented in Trial~2. Separate response headers cannot preserve chunked sessions if the read path has already divided an \texttt{Expires} attribute at its comma.

\subsubsection{EPD Evaluation without Experience Context}

\noindent\textbf{Observed behavior in the selected EPD patch.}
At evaluation, the model trained with Experience Distillation receives the original task issue and repository state but not the two-trial experience context. The selected lines below implement both protocol directions; unchanged code connecting the parser result to the writer is omitted.

\begin{sweedit}{EPD Evaluation Patch Excerpt: Reconstruct and Emit Chunked Session Cookies}
\ttfamily\scriptsize
if (!value) return [];\\
const cookieList = [];\\
let currentCookie = '';\\
for (const part of value.split(',')) \{\\
\  if (!currentCookie) \{\\
\    currentCookie = part.trim(); continue;\\
\  \}\\
\  const next = part.trimStart();\\
\  if (/\char94{}[a-z][a-z0-9-\_]*=/i.test(next)) \{\\
\    cookieList.push(currentCookie.trim());\\
\    currentCookie = next;\\
\  \} else currentCookie += ',' + part;\\
\}\\
if (currentCookie) cookieList.push(currentCookie.trim());\\
return cookieList;\\
\mbox{[\ldots]}\\
res.headers.delete(\char39{}set-cookie\char39{});\\
cookies.forEach(cookie => \{\\
\  res.headers.append(\char39{}set-cookie\char39{}, cookie);\\
\});
\end{sweedit}

\noindent\textbf{Case-level outcome.}
The selected parser uses a name/value-prefix heuristic to distinguish a new cookie from an attribute continuation, rejoins the latter, and returns the reconstructed list. The displayed write path clears the combined header and appends each reconstructed cookie. This is the concrete accepted implementation, not a claim that the regular expression defines every valid cookie-name boundary. The patch therefore exhibits both protocol directions observed under experience-conditioned ICL without receiving the two-trial experience context at evaluation. All 49 tests in the candidate's reported utility-test run pass. Among the nine EPD candidates retained for this task, eight are accepted; all ten ICL candidates are accepted, whereas zeroshot, SFT, and PPO each produce zero accepted candidates from ten samples.

\swecasehead{Case Study 5 on Curated SWE Task: Configurable GraphQL Root Type Names}

\begin{swetask}{Task Issue Summary: GraphQL Mesh Root Type Names}
\small
\textbf{Project and language.} GraphQL Mesh composes schemas and applies TypeScript transforms. Its encapsulation transform wraps root operations from one source under a generated field.

\textbf{User-visible failure.} The transform assumes that every source schema names its root operation types \texttt{Query}, \texttt{Mutation}, and \texttt{Subscription}. A schema using names such as \texttt{QueryRoot} cannot be encapsulated because the transform looks up and rewrites the wrong outer type.

\textbf{Requested behavior.} Allow users to configure the query, mutation, and subscription outer-root names used by the encapsulation transform.
\end{swetask}

\subsubsection{Accumulated Experience and Experience-Conditioned ICL}

\noindent\textbf{Repair attempts across six trials.}
The issue appears to require three parameter substitutions, but the accumulated trials show that the names participate in schema lookup, generated wrapper construction, lifecycle-dependent defaults, and public configuration.

\begin{center}
\small
\setlength{\tabcolsep}{4pt}
\begin{tabular}{@{}>{\centering\arraybackslash}p{0.10\linewidth} >{\raggedright\arraybackslash}p{0.68\linewidth} >{\centering\arraybackslash}p{0.13\linewidth}@{}}
\toprule
\textbf{Trials} & \textbf{Principal new change or revised hypothesis} & \textbf{Outcome} \\
\midrule
1--3 & Add the three public configuration fields and generated artifacts, generalize source-root lookup to string keys, and preserve stable generated names. These attempts still resolve absent options to conventional names before the source schema is available. & Rejected \\
4 & Also derive the generated inner wrapper names from the configured outer names. This couples two names that serve different roles in the transformed schema. & Rejected \\
5 & Move root resolution to schema time and detect defaults from the input schema, but still derive generated wrapper names from the resolved outer names. & Rejected \\
6 & Restore stable generated names such as \texttt{SomeSourceQuery} while retaining the configuration and schema-time source-root resolution introduced earlier. & Accepted \\
\bottomrule
\end{tabular}
\captionof{table}{Evolution of the GraphQL Mesh repair. The repeated failures separate source-schema lookup names from names created by the encapsulation transform.}
\end{center}

\begin{sweknowledge}{Task-Specific Knowledge Made Available by Repeated-Trial Experience}
\textbf{Issue-and-repository-only zeroshot behavior.} All ten sampled trajectories refer to the three explicit root-name overrides named in the issue, but only five modify a public configuration artifact and only three integrate the options into runtime behavior. None defers default resolution until the input schema is available. Of the three runtime patches, two preserve stable generated names such as \texttt{SomeSourceQuery}, whereas one derives generated names from configured source-root names. The complete separation between schema-discovered source roots and stable generated wrapper names is therefore not reliably composed under issue-and-repository-only evaluation. \textbf{Knowledge made available by repeated trials.} Trial~5 defers source-root lookup until the schema is available, but incorrectly uses the detected source name as the generated wrapper name. Trial~6 preserves the schema-time lookup while restoring stable generated names such as \texttt{SomeSourceQuery}. For example, if the source named \texttt{SomeSource} exposes a query root called \texttt{QueryRoot}, the transform must find the existing \texttt{QueryRoot} type but still create the wrapper name \texttt{SomeSourceQuery}. The accepted repair therefore separates names discovered from the source schema from wrapper names generated by GraphQL Mesh and exposes the explicit overrides through the public configuration surfaces.
\end{sweknowledge}

The ICL reference receives all six trials, and nine of ten sampled candidates are accepted. In one selected accepted patch, each source-root name is resolved from explicit configuration, then from the available schema, and finally from the conventional fallback. The generated wrapper name remains based on the source name. The query branch below shows this separation; the same patch applies the corresponding structure to mutation and subscription and adds all three options to the public configuration interfaces.

\begin{sweedit}{Selected Accepted ICL Patch Excerpt: Separate Source and Generated Root Names}
\ttfamily\scriptsize
if (this.applyTo.query) \{\\
\  const outerName = this.queryOuterType ||\\
\    originalWrappingSchema.getQueryType()?.name || 'Query';\\
\  const transform = new WrapType(\\
\    outerName, `\$\{this.name\}Query`, this.name);\\
\  \mbox{[\ldots]}\\
\}
\end{sweedit}

\subsubsection{Teacher-Generated Distillation Targets}

Conditioned on the accumulated experience, separate decisions in the selected branch-packed sequence express schema-time root resolution, stable generated naming, and the corresponding public configuration.

\begin{sweevidence}{Teacher Target 1: Resolve Defaults after the Source Schema Is Available}
\small
\swetargetlink{Schema-time fallback introduced in Trial~5.}{The first three trials resolve absent overrides before the input schema is available. Trial~5 moves this decision to schema transformation, where GraphQL Mesh can inspect the actual source-root names.}
``Move \texttt{WrapType} creation from constructor to \texttt{generateSchemaTransforms} so we can get the actual root names from the original schema if custom names are not provided''
\end{sweevidence}

\medskip
\begin{sweevidence}{Teacher Target 2: Preserve Stable Generated Wrapper Names}
\small
\swetargetlink{Decisive Trial~6 correction.}{Trial~5 obtains the correct source-root names but also uses them to derive generated wrapper names. Trial~6 separates those roles and restores the original Mesh-generated naming pattern.}
``Keep the original inner naming pattern to maintain backwards compatibility''
\end{sweevidence}

\medskip
\begin{sweevidence}{Teacher Target 3: Expose Root-Name Overrides through Public Configuration}
\small
\swetargetlink{Public configuration carried through the accepted repair.}{Trials~1--3 add the three override fields. The selected teacher sequence regenerates those fields in the YAML schema and carries them into the generated JSON schema and TypeScript interface.}
``First, add the new configuration options to \texttt{yaml-config.graphql} as requested.''

\noindent[\ldots]

``Updated \texttt{config-schema.json} with new configuration [\ldots] Updated \texttt{config.ts} TypeScript types with new optional properties''
\end{sweevidence}

\pagebreak[3]
These decisions occur at different branch points in the selected sequence. Collectively, they express the distinction established by Trials~5--6: inspect the available source schema to find existing root names, preserve GraphQL Mesh's generated naming convention, and expose explicit overrides through public configuration.

\subsubsection{EPD Evaluation without Experience Context}

\noindent\textbf{Observed behavior in the selected EPD patch.}
At evaluation, the EPD-trained model receives the original task issue and repository state, but not the six-trial experience context. The runtime excerpt implements the accepted distinction completed in Trial~6: configuration or the available schema determines which existing root type is wrapped, while GraphQL Mesh continues to generate stable names from the source name.

\begin{sweedit}{EPD Evaluation Patch Excerpt: Root Resolution in \texttt{encapsulate/src/index.ts}}
\ttfamily\scriptsize
\textnormal{\textit{Constructor configuration:}}\\
this.customQueryOuterType = config?.queryOuterType;\\
this.customMutationOuterType = config?.mutationOuterType;\\
this.customSubscriptionOuterType = config?.subscriptionOuterType;\\
\mbox{[\ldots]}\\
\textnormal{\textit{Query branch in \texttt{generateSchemaTransforms}:}}\\
if (this.applyTo.query) \{\\
\  const queryOuterType = this.customQueryOuterType ||\\
\    originalWrappingSchema.getQueryType()?.name || 'Query';\\
\  const transformedName = `\$\{this.name\}Query`;\\
\  const transform = new WrapType(\\
\    queryOuterType, transformedName, this.name);\\
\  \mbox{[\ldots]}\\
\}\\
\mbox{[\ldots]}\\
\textnormal{\textit{Mutation branch in \texttt{generateSchemaTransforms}:}}\\
if (this.applyTo.mutation) \{\\
\  const mutationOuterType = this.customMutationOuterType ||\\
\    originalWrappingSchema.getMutationType()?.name || 'Mutation';\\
\  const transformedName = `\$\{this.name\}Mutation`;\\
\  const transform = new WrapType(\\
\    mutationOuterType, transformedName, this.name);\\
\  \mbox{[\ldots]}\\
\}\\
\mbox{[\ldots]}\\
\textnormal{\textit{Subscription branch in \texttt{generateSchemaTransforms}:}}\\
if (this.applyTo.subscription) \{\\
\  const subscriptionOuterType = this.customSubscriptionOuterType ||\\
\    originalWrappingSchema.getSubscriptionType()?.name || 'Subscription';\\
\  const transformedName = `\$\{this.name\}Subscription`;\\
\  const transform = new WrapType(\\
\    subscriptionOuterType, transformedName, this.name);\\
\  \mbox{[\ldots]}\\
\}
\end{sweedit}

The public-configuration excerpt shows the three override fields first introduced in Trials~1--3. The selected diff adds the same options to the transform's YAML schema, the TypeScript configuration interface, and the generated JSON schema.

\begin{sweedit}{EPD Evaluation Patch Excerpt: Root-Name Options in Public Configuration}
\ttfamily\scriptsize
\textnormal{\textit{\texttt{yaml-config.graphql}:}}\\
queryOuterType: String\\
mutationOuterType: String\\
subscriptionOuterType: String\\
\mbox{[\ldots]}\\
\textnormal{\textit{\texttt{packages/types/src/config.ts}:}}\\
queryOuterType?: string;\\
mutationOuterType?: string;\\
subscriptionOuterType?: string;\\
\mbox{[\ldots]}\\
\textnormal{\textit{\texttt{packages/types/src/config-schema.json}:}}\\
\char34{}queryOuterType\char34{}: \{ \char34{}type\char34{}: \char34{}string\char34{} \}\\
\mbox{[analogous mutation and subscription fields]}
\end{sweedit}

\noindent\textbf{Case-level outcome.}
The selected EPD patch exhibits the outer-versus-generated naming distinction identified through experience-conditioned ICL for query, mutation, and subscription roots. It also exposes the corresponding public configuration fields. Its tests report 146 passed and one skipped. Nine of ten EPD candidates and nine of ten ICL candidates are accepted, compared with zero of ten candidates from each of zeroshot, SFT, and PPO.

\swecasehead{Case Study 6 on Curated SWE Task: Squash-Merge Commit Message Selection}

\begin{swetask}{Task Issue Summary: GitHub CLI Squash-Merge Messages}
\small
\textbf{Project and language.} GitHub CLI is a Go command-line client. The task concerns the behavior of \texttt{gh pr merge} for squash merges.

\textbf{User-visible failure.} For squash merges, the existing command derives the resulting commit message from the last commit, which may be an uninformative message such as a lint or rebase update. The issue requests a choice among the pull-request title, the last commit message, and a custom message.

\textbf{Requested behavior.} Preview the squash-merge commit message in the CLI and let the user choose the pull-request title, the last commit message, or a custom message.
\end{swetask}

\subsubsection{Accumulated Experience and Experience-Conditioned ICL}

\noindent\textbf{Repair attempts across five trials.}
The requested choice appears local to a terminal prompt, but each attempt exposes another part of the command-to-API path.

\begin{center}
\small
\setlength{\tabcolsep}{4pt}
\begin{tabular}{@{}>{\centering\arraybackslash}p{0.08\linewidth} >{\raggedright\arraybackslash}p{0.70\linewidth} >{\centering\arraybackslash}p{0.13\linewidth}@{}}
\toprule
\textbf{Trial} & \textbf{Principal new change or revised hypothesis} & \textbf{Outcome} \\
\midrule
1 & Add a three-way squash-message prompt and extend the merge mutation to accept a selected headline and body. Prompting enters command paths that previously completed without input. & Rejected \\
2 & Restrict prompting to fewer flows so existing command tests complete, but leave the interaction condition and headless default unresolved. & Rejected \\
3 & Prompt whenever a squash merge runs on a TTY, including explicit \texttt{--squash}. This still treats terminal availability as equivalent to interactive intent. & Rejected \\
4 & Establish pull-request title/body as the deterministic squash default, but still gate the prompt on TTY alone. & Rejected \\
5 & Require both interactive mode and stdin TTY before prompting, while retaining the query data, mutation fields, and headless default introduced earlier. & Accepted \\
\bottomrule
\end{tabular}
\captionof{table}{Evolution of the GitHub CLI repair. The trials expand a terminal prompt into an explicit contract for interactive and headless command execution.}
\end{center}

\begin{sweknowledge}{Task-Specific Knowledge Made Available by Repeated-Trial Experience}
\textbf{Issue-and-repository-only zeroshot behavior.} All ten sampled trajectories locate the merge command and trace its call into \texttt{api.PullRequestMerge}, but every trajectory terminates during repository inspection without applying an edit. Consequently, none implements the query extension, pull-request-title/body default, interaction gate, three-way selection, or mutation propagation that appear together in the accepted repair. Under issue-and-repository-only evaluation, the complete command-to-API behavior is therefore not reliably composed. \textbf{Knowledge made available by repeated trials.} The repeated trials establish the title/body default before prompting, restrict the prompt to interactive mode when stdin is attached to a TTY, fetch the last commit for the corresponding option, and pass the selected fields to the merge mutation.
\end{sweknowledge}

The ICL reference receives all five trials. One selected accepted repair separates interactive and headless behavior, obtains the data needed for every offered option, and carries the selected message through the GraphQL mutation. All ten ICL candidates are accepted under the pass@10 budget.

\subsubsection{Teacher-Generated Distillation Targets}

Conditioned on the accumulated experience, the selected teacher targets retain Trial~1's query and API path, Trial~4's deterministic headless default, and Trial~5's corrected interaction gate.

\begin{sweevidence}{Teacher Target 1: Use the Pull-Request Message as the Headless Default}
\small
\swetargetlink{Accepted Trial~4 component: deterministic noninteractive behavior.}{Trials~1--3 expose command paths that must not block for input. Trial~4 establishes the pull-request title and body as the squash-message default, which Trial~5 retains.}
``set the default commit message for squash merges to be the PR title + PR body (fixing the core issue)''
\end{sweevidence}

\medskip
\begin{sweevidence}{Teacher Target 2: Gate the Three-Way Prompt on Interactive Mode and TTY}
\small
\swetargetlink{Decisive Trial~5 correction.}{Trial~1 introduces the three choices, but Trials~2--4 do not yet distinguish terminal availability from interactive intent. Trial~5 requires both interactive mode and stdin attached to a TTY before presenting the prompt.}
``Add the interactive prompt when in interactive mode and TTY to allow users to select between the three options: PR title, last commit, custom message''
\end{sweevidence}

\medskip
\begin{sweevidence}{Teacher Target 3: Fetch and Propagate the Selected Message}
\small
\swetargetlink{Accepted Trial~1 data-and-API component retained through Trial~5.}{Trial~1 extends the pull-request query and merge mutation so that the three-way choice can read the last commit and transmit the selected headline and body. Trials~2--5 revise when prompting occurs and what the default should be, while retaining this cross-layer path.}
``Update all GraphQL queries that fetch commits to get the last commit's message''

\noindent[\ldots]

``Update PullRequestMerge function to accept commitHeadline and commitBody parameters''

\noindent[\ldots]

``Modify the mutation to add CommitHeadline and CommitBody when they're not empty''

\noindent[\ldots]

``Update all the calls to api.PullRequestMerge to include the new parameters''
\end{sweevidence}

Together, the three targets cover the deterministic default, the interactive override, and the query-to-mutation data path combined by the accepted trial.

\subsubsection{EPD Evaluation without Experience Context}

\noindent\textbf{Observed behavior in the selected EPD patch.}
At evaluation, the EPD-trained model receives the original task issue and repository state, but not the five-trial experience context. The first excerpt adds a \texttt{Message} field to each queried commit node and requests it from the last commit.

\begin{sweedit}{EPD Evaluation Patch Excerpt: Fetch the Last Commit Message}
\ttfamily\scriptsize
Message string\\
\mbox{[\ldots]}\\
commits(last: 1) \{\\
\  nodes \{\\
\    commit \{ message \}\\
\  \}\\
\}
\end{sweedit}

The next excerpt shows how the accepted repair combines Trial~4's deterministic headless default with Trial~5's interaction gate. The two nondefault choices replace the headline and body with the last commit or a custom message.

\begin{sweedit}{EPD Evaluation Patch Excerpt: Select the Squash-Merge Message}
\ttfamily\scriptsize
prTitle := fmt.Sprintf(\char34{}\%s (\#\%d)\char34{}, pr.Title, pr.Number)\\
if mergeMethod == api.PullRequestMergeMethodSquash \{\\
\  commitHeadline = prTitle; commitBody = pr.Body\\
\}\\
\mbox{[\ldots]}\\
if mergeMethod == api.PullRequestMergeMethodSquash \&\&\\
\  opts.InteractiveMode \&\& opts.IO.IsStdinTTY() \{\\
\  lastCommitMessage = strings.TrimSpace(\\
\    pr.Commits.Nodes[len(pr.Commits.Nodes)-1].Commit.Message)\\
\  options := []string\{\\
\    fmt.Sprintf(\char34{}Use pull request title: \%q\char34{}, prTitle),\\
\    fmt.Sprintf(\char34{}Use last commit message: \%q\char34{}, lastCommitMessage),\\
\    \char34{}Enter custom commit message\char34{},\\
\  \}\\
\  \mbox{[\ldots]}\\
\  switch(choice) \{\\
\  case 1:\\
\    if lines := strings.SplitN(lastCommitMessage,\\
\      \char34{}\textbackslash n\char34{}, 2); len(lines) >= 2 \{\\
\      commitHeadline = lines[0]\\
\      commitBody = strings.TrimSpace(lines[1])\\
\    \} else \{ \mbox{[\ldots]} \}\\
\  case 2:\\
\    \mbox{[\ldots] SurveyAskOne(editorQuestion, \&customMessage)}\\
\    if lines := strings.SplitN(strings.TrimSpace(customMessage),\\
\      \char34{}\textbackslash n\char34{}, 2); len(lines) >= 2 \{\\
\      commitHeadline = lines[0]\\
\      commitBody = strings.TrimSpace(lines[1])\\
\    \} else \{ \mbox{[\ldots]} \}\\
\  \}\\
\}\\
\end{sweedit}

The final excerpt shows that the selected values cross the GraphQL API boundary rather than remaining local command state.

\begin{sweedit}{EPD Evaluation Patch Excerpt: Pass the Message to the Merge Mutation}
\ttfamily\scriptsize
if commitHeadline != \char34{}\char34{} \{\\
\  commitHeadline := githubv4.String(commitHeadline)\\
\  input.CommitHeadline = \&commitHeadline\\
\}\\
if commitBody != \char34{}\char34{} \{\\
\  commitBody := githubv4.String(commitBody)\\
\  input.CommitBody = \&commitBody\\
\}
\end{sweedit}

\noindent\textbf{Case-level outcome.}
The selected EPD patch implements the headless default, interactive override, and query-to-mutation path identified across the five trials. It exhibits this command/API contract without receiving the experience context at evaluation. The squash, non-TTY, and interactive merge-command tests pass. Nine of ten EPD candidates are accepted, compared with ten of ten ICL candidates and zero of ten candidates from each of zeroshot, SFT, and PPO.

\swecasehead{Case Study 7 on Curated SWE Task: Stale Duplicate Error after Circular-Reference Recovery}

\begin{swetask}{Task Issue Summary: Gauge Circular-Reference Recovery}
\small
\textbf{Project and language.} Gauge is a Go test-automation framework. A \texttt{.cpt} file defines reusable \emph{concepts}, each expanding to a sequence of test steps.

\textbf{User-visible failure.} After a user creates and then removes a circular reference in a \texttt{.cpt} file, Gauge continues to report \texttt{Duplicate concept definition found}; the report provides no additional logs.

\textbf{Expected behavior.} Removing the circular reference should also remove the stale duplicate-definition error.
\end{swetask}

\subsubsection{Accumulated Experience and Experience-Conditioned ICL}

\noindent\textbf{Repair attempts across two trials.}
Trial~1 modifies both the editor caller and parser cleanup. Trial~2 retains those edits and moves file deduplication to the shared discovery utilities used outside the editor.

\begin{center}
\small
\setlength{\tabcolsep}{4pt}
\begin{tabular}{@{}>{\centering\arraybackslash}p{0.08\linewidth} >{\raggedright\arraybackslash}p{0.70\linewidth} >{\centering\arraybackslash}p{0.13\linewidth}@{}}
\toprule
\textbf{Trial} & \textbf{Principal new change or revised hypothesis} & \textbf{Outcome} \\
\midrule
1 & Deduplicate concept-file inputs in the language-server caller and replace direct \texttt{ConceptsMap} deletion with \texttt{conceptDictionary.Remove}, which clears both parser maps. Shared non-editor callers can still receive duplicate files. & Rejected \\
2 & Retain dictionary-level cleanup and move concept/specification-file deduplication into the shared discovery utilities used by all callers. & Accepted \\
\bottomrule
\end{tabular}
\captionof{table}{Evolution of the Gauge repair. Trial~2 broadens caller-local deduplication into a shared file-discovery invariant while retaining complete parser-state cleanup.}
\end{center}

\begin{sweknowledge}{Task-Specific Knowledge Made Available by Repeated-Trial Experience}
\textbf{Issue-and-repository-only zeroshot behavior.} All ten sampled trajectories terminate during parser and file-discovery inspection without applying an edit. One explicitly diagnoses that circular-reference cleanup should call \texttt{conceptDictionary.Remove} rather than delete only from \texttt{ConceptsMap}, but none implements that change; none adds either caller-local or shared file deduplication. The complete combination of two-map cleanup and shared deduplication is therefore not reliably produced under issue-and-repository-only evaluation. \textbf{Knowledge made available by repeated trials.} Trial~1 combines caller-local deduplication with complete dictionary removal, but remains rejected. Trial~2 retains the parser cleanup and moves deduplication into shared \texttt{GetConceptFiles} and \texttt{GetSpecFiles}. Dictionary-level removal prevents a failed concept from remaining in the secondary \texttt{constructionMap}, while shared deduplication prevents the same concept file from re-entering the parser through repeated discovery paths. The accepted state therefore removes failed concepts from every internal representation and makes discovery idempotent across callers while leaving genuine duplicate definitions reportable.
\end{sweknowledge}

The ICL reference receives both trials. One selected accepted repair uses the dictionary's complete removal operation and deduplicates files before parsing, so the same behavior applies to CLI and editor paths. All ten ICL candidates are accepted under the pass@10 budget.

\subsubsection{Teacher-Generated Distillation Targets}

Conditioned on the accumulated experience, the teacher identifies shared file discovery as the missing scope change and separately retains dictionary-level cleanup. Its first quoted diagnosis overlooks the dictionary cleanup already added in Trial~1, but the generated edit matches the accepted Trial~2 change from caller-local to shared deduplication.

\begin{sweevidencecompact}{Teacher Target 1: Move Deduplication into Shared File Discovery}
\small
\swetargetlink{Accepted Trial~2 scope correction; the diagnosis below partially misstates Trial~1.}{Trial~1 already changes parser cleanup and deduplicates in the language-server caller. Trial~2 applies \texttt{uniqueStrings} in the shared \texttt{GetConceptFiles} and \texttt{GetSpecFiles} utilities.}
``So my previous fix in trial 1 only fixed diagnostics (LSP/editor), but didn't fix the core GetConceptFiles. The problem is that the bug can also happen on CLI when running tests, not just in the editor.''

\noindent[\ldots]

``Added deduplication to both \texttt{GetSpecFiles} and \texttt{GetConceptFiles} using a new helper function \texttt{uniqueStrings}.''
\end{sweevidencecompact}

\medskip
\begin{sweevidencecompact}{Teacher Target 2: Remove Circular-Reference State from Both Parser Maps}
\small
\swetargetlink{Accepted repair component retained from Trial~1.}{Trial~1 already replaces direct deletion from \texttt{ConceptsMap} with \texttt{conceptDictionary.Remove}; Trial~2 retains this change while correcting deduplication scope. The quote below accurately states the two-map cleanup performed by that operation.}

``Changed the circular reference cleanup from just deleting from \texttt{ConceptsMap} to using \texttt{conceptDictionary.Remove()} which correctly cleans up both \texttt{ConceptsMap} and \texttt{constructionMap}.''
\end{sweevidencecompact}

Together, the two targets specify the two accepted repair components. The EPD patch below omits Trial~1's now-redundant language-server-local deduplication.

\subsubsection{EPD Evaluation without Experience Context}

\noindent\textbf{Observed behavior in the selected EPD patch.}
At evaluation, the EPD-trained model receives the original task issue and repository state, but not the two-trial experience context. The selected patch contains both shared operations identified above:

\begin{sweedit}{EPD Evaluation Patch Excerpt: Shared Duplicate-Concept Recovery}
\ttfamily\small
\textnormal{\textit{\texttt{ValidateConcepts}:}}\\
conceptDictionary.Remove(concept.ConceptStep.Value)\\
\mbox{[\ldots]}\\
\textnormal{\textit{\texttt{GetSpecFiles}:}}\\
return uniqueStrings(specFiles)\\
\mbox{[\ldots]}\\
\textnormal{\textit{\texttt{GetConceptFiles}:}}\\
return uniqueStrings(files)
\end{sweedit}

\noindent\textbf{Case-level outcome.}
The selected EPD patch removes failed concepts through the dictionary abstraction, which updates both parser maps, and applies \texttt{uniqueStrings} to both discovery paths. The patch exhibits the parser-state and input-idempotence behavior observed under experience-conditioned ICL without receiving the two-trial experience context at evaluation. The parser reports 198 passing tests; the utility and Gauge suites also pass, including checks that intentional duplicates remain detected. Nine of ten EPD candidates are accepted, compared with ten of ten ICL candidates and zero of ten candidates from each of zeroshot, SFT, and PPO.

\swecasehead{Case Study 8 on Curated SWE Task: Backend-Relative Stored Media URLs in Strapi}

\begin{swetask}{Task Issue Summary: Strapi Backend-Relative Media URLs}
\small
\textbf{Project and language.} Strapi is a JavaScript/TypeScript content platform. The report concerns its media-library administration interface.

\textbf{User-visible failure.} Opening the media library produces a white page with \texttt{TypeError: Failed to construct 'URL': Invalid URL}. The reporter notes that the stored \texttt{files.url} values all begin with a slash and identifies this data characteristic as the apparent trigger.

\textbf{Expected behavior.} The media-library page should open when existing records contain slash-prefixed media URLs.
\end{swetask}

\subsubsection{Accumulated Experience and Experience-Conditioned ICL}

\noindent\textbf{Repair attempts across four trials.}
The repeated trials explore stored-media parsing, add defensive provider handling, and eventually identify Strapi's backend-prefix helper as the repository-specific mechanism for stored paths. The accepted fourth trial still broadens the add-from-URL schema; the teacher later restores strict validation for user-entered URLs.

\begin{center}
\small
\setlength{\tabcolsep}{4pt}
\begin{tabular}{@{}>{\centering\arraybackslash}p{0.08\linewidth} >{\raggedright\arraybackslash}p{0.70\linewidth} >{\centering\arraybackslash}p{0.13\linewidth}@{}}
\toprule
\textbf{Trial} & \textbf{Principal new change or revised hypothesis} & \textbf{Outcome} \\
\midrule
1 & Supply a browser base to frontend URL parsers, including the ``add from URL'' schema. Relative stored paths become parseable, but user-entered relative URLs also become valid. & Rejected \\
2 & Retain the frontend changes and add defensive handling in the storage-provider URL utilities. The repair still applies one policy to distinct data sources. & Rejected \\
3 & Add guarded frontend fallbacks while keeping the user-input schema permissive toward relative values. & Rejected \\
4 & Replace ad hoc bases with Strapi's backend-prefix helper for stored paths and retain defensive provider handling. The patch fixes the reported failure but still applies the prefix helper inside the user-input schema. & Accepted \\
\bottomrule
\end{tabular}
\captionof{table}{Evolution of the Strapi repair. The trials establish stored-path normalization but leave an overbroad user-input schema change for the teacher to correct.}
\end{center}

\begin{sweknowledge}{Task-Specific Knowledge Made Available by Repeated-Trial Experience}
\textbf{Issue-and-repository-only zeroshot behavior.} All ten sampled trajectories locate the one-argument \texttt{new URL(...)} sites relevant to slash-prefixed paths, but five terminate before applying an edit. The other five add a browser-derived base in \texttt{appendSearchParamsToUrl}; four also change filename extraction, three also relax the user-input schema, and one adds local fallbacks. None uses Strapi's \texttt{prefixFileUrlWithBackendUrl} helper or edits the provider utilities. Thus none combines the repository's backend-prefix helper in the reported media-library path with defensive provider handling; three edited candidates also broaden user-input validation. \textbf{Knowledge made available by repeated trials.} Trial~4 replaces browser-origin resolution with the repository-specific backend-prefix helper across the affected stored-media utilities and retains defensive provider handling. The same accepted trial still relaxes the ``add from URL'' schema, so preserving strict validation for user-entered URLs is a later teacher refinement rather than knowledge established by the repeated-trial outcome.
\end{sweknowledge}

The ICL reference receives all four trials. One selected accepted repair applies the existing prefix helper to stored media paths and adds defensive provider handling; like Trial~4, it also relaxes the user-input schema. All ten ICL candidates are accepted under the pass@10 budget. The teacher sequence retains the trial-derived stored-path and provider changes and independently restores the original user-input policy.

\subsubsection{Teacher-Generated Distillation Targets}

Conditioned on the task-specific accumulated experience, the teacher separates the stored-path normalization reached in Trial~4 from the schema policy that Trial~4 changed unnecessarily.

\begin{sweevidence}{Teacher Target 1: Apply the Backend Prefix before URL Parsing}
\small
\swetargetlink{Accepted stored-path repair component.}{Trials~1--3 add browser-derived bases or local fallbacks. Trial~4 identifies Strapi's existing backend-prefix helper; the teacher explicitly reuses that repository abstraction before parsing a potentially relative path.}
``the function is called with a URL that might still be relative [\ldots] every other function in the codebase that needs an absolute file URL handles it by calling \texttt{prefixFileUrlWithBackendUrl}''
\end{sweevidence}

\medskip
\begin{sweevidence}{Teacher Target 2: Restore Absolute-URL Validation for User Input}
\small
\swetargetlink{Teacher revision of Trial~4's accepted patch.}{Trial~4 also invokes the backend-prefix helper from the ``add from URL'' schema, making relative user input valid. The teacher distinguishes that form from stored media paths and restores the schema's original absolute-URL policy.}
``\texttt{urlYupSchema.js} is already back to its original state, which is correct. We shouldn't modify it at all because it's meant to validate that user-input URLs are already absolute URLs.''
\end{sweevidence}

\medskip
\begin{sweevidence}{Teacher Target 3: Guard Provider-Side URL Parsing}
\small
\swetargetlink{Defensive behavior retained from the repeated trials.}{Trials~2--4 add provider-side handling so malformed or relative values cannot produce an unhandled server-side parse failure. The selected teacher sequence explicitly revisits this provider path.}
``Now let me also add the defensive error handling to the AWS S3 provider utils''
\end{sweevidence}

Together, the three targets separate backend-relative stored paths from user-entered external URLs and retain defensive provider behavior: normalize stored paths with the repository's backend-prefix helper, preserve absolute-only validation for user input, and guard provider-side parsing.

\subsubsection{EPD Evaluation without Experience Context}

\noindent\textbf{Observed behavior in the selected EPD patch.}
At evaluation, the model trained with Experience Distillation receives the original task issue and repository state but not the four-trial experience context. The complete selected patch leaves \texttt{urlYupSchema.js} unchanged. The first excerpt shows the reported media-library fix in \texttt{appendSearchParamsToUrl} and defensive reuse of the same helper in filename extraction.

\begin{sweedit}{EPD Evaluation Patch Excerpt: Prefix Stored Media Paths before Parsing}
\ttfamily\small
\textnormal{\textit{\texttt{appendSearchParamsToUrl}:}}\\
const prefixedUrl =\\
\  prefixFileUrlWithBackendUrl(url);\\
const urlObj = new URL(prefixedUrl);\\
\mbox{[\ldots]}\\
\textnormal{\textit{\texttt{getFilenameFromURL}:}}\\
function getFilenameFromURL(url) \{\\
\  const prefixedUrl = prefixFileUrlWithBackendUrl(url);\\
\  return new URL(prefixedUrl).pathname.split(\char39{}/\char39{}).pop();\\
\}
\end{sweedit}

The provider-side excerpt shows the complementary behavior: an invalid provider URL no longer produces an unhandled parse failure.

\enlargethispage{\baselineskip}
\begin{sweedit}{EPD Evaluation Patch Excerpt: Guard Provider-Side URL Parsing}
\ttfamily\small
try \{\\
\  url = new URL(fileUrl);\\
\} catch (err) \{\\
\  return false;\\
\}
\end{sweedit}

\noindent\textbf{Case-level outcome.}
The helper use in \texttt{appendSearchParamsToUrl} and the provider guard correspond to trial-derived behavior; leaving \texttt{urlYupSchema.js} unchanged corresponds to the teacher's later refinement. The patch also reuses the helper for filename extraction. Nine of ten EPD candidates are accepted without the four-trial experience context at evaluation, compared with ten of ten ICL candidates and zero of ten candidates from each of zeroshot, SFT, and PPO.

\subsection{Cross-Case Findings and Limitations}

Across the eight cases, the same base model exhibits different repair behavior when conditioned on repeated-trial experience. The ten issue-and-repository-only samples for each case do not produce an accepted repair, whereas the experience-conditioned ICL samples implement the task-specific behavior summarized in Table~\ref{tab:swe-cross-case-retention}. Selected teacher targets encode all or part of this behavior. A selected sequence can also revisit a rejected collection hypothesis, as in MapStore2, or refine an accepted collection patch, as in Strapi; the Billboard.js sequence does not regenerate every edit in the accepted working tree. The selected EPD patches implement the summarized behavior at evaluation without the experience context.

\begin{center}
\footnotesize
\setlength{\tabcolsep}{3pt}
\begin{tabular}{@{}>{\raggedright\arraybackslash}p{0.13\linewidth} >{\raggedright\arraybackslash}p{0.23\linewidth} >{\raggedright\arraybackslash}p{0.28\linewidth} >{\raggedright\arraybackslash}p{0.26\linewidth}@{}}
\toprule
\textbf{Case} & \textbf{Behavior across ten issue-and-repository-only zeroshot samples} & \textbf{Behavior available through experience-conditioned ICL} & \textbf{Behavior observed in selected EPD patch} \\
\midrule
Geometry reset & Local filter and state edits; no complete reset-to-refresh path. & Emit an explicit null reset and preserve it through query refresh. & Emits \texttt{value: null} and preserves the geometry update in the query epic. \\
Tooltip alignment & Nearest-point lookup and local guards; no coordinated cross-layer semantics. & Combine representation-aware lookup with absence handling across propagation, API, and DOM layers. & Uses shared-x binary search, a multiple-x fallback, and explicit absence guards. \\
Subscription cleanup & Garbage-collection or unsubscribe edits that conflate record and subscription lifetimes. & Separate cache-retained records from active live subscriptions during store collection. & Tracks retained and active references separately and unsubscribes their difference when garbage collection runs. \\
Session cookies & Emission changes and partial comma handling; no complete parser--writer protocol. & Reconstruct cookie boundaries while preserving attribute commas, then emit separate headers. & Detects cookie boundaries, preserves \texttt{Expires} fragments, and appends separate headers. \\
GraphQL roots & All ten reference explicit overrides; five edit public configuration, three integrate runtime use, and none resolves absent overrides from the input schema. & Separate schema-discovered source roots from stable generated wrapper names. & Resolves source roots at schema time and preserves generated names across configuration surfaces. \\
Squash messages & All ten trace the merge command into the API helper but stop before editing; no query, default, prompt-gating, or mutation change is implemented. & Combine a deterministic headless default, gated interactive choices, and API propagation. & Supplies the default, gates prompts on interactive mode and TTY, and passes mutation fields. \\
Parser recovery & All ten stop before editing; one diagnoses two-map cleanup, and none implements caller-local or shared deduplication. & Combine complete parser cleanup with shared file-discovery deduplication. & Uses dictionary-level removal and deduplicates both shared discovery paths. \\
Media URLs & Five add browser-based handling in an admin URL utility, three also relax input validation, and none uses the backend-prefix helper or adds provider guards. & Normalize stored paths with the backend helper and retain provider-side guards. & Prefixes stored paths, adds defensive provider parsing, and preserves strict input validation after teacher refinement. \\
\bottomrule
\end{tabular}
\captionof{table}{Behavioral comparison across eight cases. The zeroshot column audits all ten issue-and-repository-only samples; the ICL and EPD columns summarize accepted experience-conditioned repairs and one selected EPD patch generated without the experience context.}
\label{tab:swe-cross-case-retention}
\end{center}

\noindent\textbf{Outcome contrast across the selected cases.}
Under the pass@10 budget, experience-conditioned ICL solves all eight cases with 79 accepted candidates out of 80, whereas zeroshot solves none. EPD also solves all eight without the experience context, yielding 72 accepted candidates out of 79 valid samples; SFT and PPO yield no accepted candidates in 80 samples per method. These counts establish the case outcomes, while the selected teacher decisions and EPD patches show the task-specific behavior observed before and after distillation.

\noindent\textbf{Scope of the qualitative evidence.}
Each teacher excerpt and EPD patch is one selected qualitative sample. Their correspondence documents an observed repair pattern but does not identify a unique causal path or estimate how frequently EPD uses that strategy. Because the cases are selected for outcome contrast, aggregate pass@1 in Table~\ref{tab:main-results} remains the evidence for overall performance.

\endgroup

\end{document}